%% file: offline_arxiv.tex
\documentclass[12pt]{article}
\usepackage[margin=1.25in]{geometry}

\usepackage[utf8]{inputenc} 
\usepackage[T1]{fontenc}    
\usepackage{hyperref}       
\usepackage{url}            
\usepackage{booktabs}       
\usepackage{amsfonts}       
\usepackage{nicefrac}       
\usepackage{microtype}      
\usepackage{xcolor}         

\usepackage{times}
\usepackage{url}            %
\usepackage{booktabs}       %
\usepackage{amsfonts}       %
\usepackage{amssymb}
\usepackage{nicefrac}       %
\usepackage{microtype}      %
\usepackage{mkolar_definitions}
\usepackage{xspace}
\usepackage{multirow}
\usepackage{amsmath}
\usepackage{algorithm}
\usepackage{algpseudocode}
\usepackage{color}
\usepackage{authblk}
\usepackage{color, colortbl}
\usepackage{enumitem}
\usepackage{comment}
\usepackage{bm}
\usepackage{subcaption}
\usepackage{bbm}
\usepackage{thmtools}

\usepackage{url}
\usepackage{authblk}
\usepackage{csquotes}
\usepackage{tikz}
\usepackage{natbib}
\usepackage[compact]{titlesec}
\allowdisplaybreaks

\newtheorem{assumption}{Assumption}

\usepackage{centernot}

\newcount\Comments  %
\definecolor{darkgreen}{rgb}{0,0.5,0}
\definecolor{darkred}{rgb}{0.7,0,0}
\definecolor{teal}{rgb}{0.3,0.8,0.8}
\definecolor{orange}{rgb}{1.0,0.5,0.0}
\definecolor{purple}{rgb}{0.8,0.0,0.8}
\newcommand{\kibitz}[2]{\ifnum\Comments=1{\textcolor{#1}{\textsf{\footnotesize #2}}}\fi}
\usetikzlibrary{positioning}

\definecolor{Gray}{gray}{0.9}

\input{notation.tex}
\newcommand{\KL}{\boldsymbol{\mathrm{KL}}}

\newcommand{\hr}{\widehat{r}}
\newcommand{\rmax}{r_{\mathrm{max}}}
\newcommand{\rstar}{r^{\star}}
\newcommand{\E}{\mathbb{E}}

\newcommand{\pis}{\pi^{\star}}
\newcommand{\VS}{V^{\star}}
\newcommand{\QS}{Q^{\star}}
\newcommand{\hpi}{\widehat{\pi}}
\newcommand{\eps}{\epsilon}
\newcommand{\Pihis}{\Pi_{\mathrm{his}}}
\newcommand{\Pimar}{\Pi_{\mathrm{Mar}}}
\newcommand{\Pstar}{P^{\star}}
\renewcommand{\rho}{P_0}
\newcommand{\rhos}{\rho^{\star}}
\newcommand{\mainalg}{\texttt{FREEHAND}\xspace}
\newcommand{\unalg}{\texttt{FREEHAND-transition}\xspace}
\newcommand{\actalg}{\texttt{FREEHAND-action}\xspace}
\newcommand{\muref}{\mu_{\mathrm{ref}}}
\newcommand{\pit}{\pi_{\mathrm{tar}}}
\newcommand{\cmle}{c_{\mathrm{MLE}}}
\newcommand{\ctv}{c_{\mathrm{TV}}}
\newcommand{\rinf}{r^{\mathrm{inf}}}
\newcommand{\TO}{\widetilde{\mathcal{O}}}
\newcommand{\Ctr}{C_{\mathrm{tr}}}
\newcommand{\Cst}{C_{\mathrm{st}}}
\newcommand{\Unif}{\mathrm{Unif}}
\newcommand{\Bern}{\mathrm{Bern}}
\newcommand{\hP}{\widehat{P}}
\newcommand{\hrho}{\widehat{\rho}}
\newcommand{\Pini}{\mathcal{P}_{\mathrm{ini}}}
\newcommand{\Pinf}{P^{\mathrm{inf}}}

\newcommand{\idx}{\mathbbm{1}}
\newcommand{\Cact}{C_{\mathrm{act}}}
\newcommand{\of}{\overline{f}}
\newcommand{\AS}{A^{\star}}
\newcommand{\hA}{\widehat{A}}

\newcommand{\ie}{\emph{i.e.}}
\newcommand{\eg}{\emph{e.g.}}

\newcommand{\BF}{\overline{\mathcal{F}}}
\newcommand{\TF}{\widetilde{\mathcal{F}}}

\ifdefined\usebigfont

\usepackage{times}
\usepackage[fontsize=13pt]{scrextend}
\AtBeginDocument{\newgeometry{letterpaper,left=1.56in,right=1.56in,top=1.71in,bottom=1.77in}}

\else
\fi

\usepackage{scrtime}
\usepackage[dont-mess-around]{fnpct}

\usepackage{etoolbox}
\makeatletter
\patchcmd\maketitle{\setcounter{footnote}{0}}{}{}{}
\patchcmd\maketitle{%
  \renewcommand\thefootnote{\@fnsymbol\c@footnote}}{\AdaptNote\thanks\multthanks}{}{}
\patchcmd\maketitle{%
  \def\@makefnmark{\rlap{\@textsuperscript{\normalfont\@thefnmark}}}}{}{}{}
\makeatother

\begin{document}
\title{Provable Offline Preference-Based Reinforcement Learning}
\author{
	Wenhao Zhan \footnote{The first two authors contributed equally.} \textsuperscript{ 
 }\thanks{Princeton University. Email: \texttt{wenhao.zhan@princeton.edu}} \quad Masatoshi Uehara \footnotemark[1] \textsuperscript{ }\thanks{Cornell University. Email: \texttt{mu223@cornell.edu}} \quad Nathan Kallus\thanks{Cornell University. Email: \texttt{kallus@cornell.edu}} \\
    Jason D. Lee\thanks{Princeton University. Email: \texttt{jasonlee@princeton.edu}} \qquad\qquad Wen Sun\thanks{Cornell University. Email: \texttt{ws455@cornell.edu}}
}

\maketitle
\begin{abstract}
In this paper, we investigate the problem of offline Preference-based Reinforcement Learning (PbRL) with human feedback where feedback is available in the form of preference between trajectory pairs rather than explicit rewards. Our proposed algorithm consists of two main steps: (1) estimate the implicit reward using Maximum Likelihood Estimation (MLE) with general function approximation from offline data and (2) solve a distributionally robust planning problem over a confidence set around the MLE. We consider the general reward setting where the reward can be defined over the whole trajectory and provide a novel guarantee that allows us to learn any target policy with a polynomial number of samples, as long as the target policy is covered by the offline data. This guarantee is the first of its kind with general function approximation. To measure the coverage of the target policy, we introduce a new single-policy concentrability coefficient, which can be upper bounded by the per-trajectory concentrability coefficient. We also establish lower bounds that highlight the necessity of such concentrability and the difference from standard RL, where state-action-wise rewards are directly observed. We further extend and analyze our algorithm when the feedback is given over action pairs.
\end{abstract}

\input{introduction}
\input{preliminary}
\input{known}
\input{unknown}
\input{action}
\input{summary}
\clearpage 
\bibliographystyle{apalike}
\bibliography{ref.bib,ref_rlhf.bib}
\clearpage
\allowdisplaybreaks
\appendix
\input{discuss}

\input{appendix_bracket}
\input{sketch}

\input{proof_lower}

\input{proof_unknown}
\input{proof_action}

\end{document}

%% file: introduction.tex
\section{Introduction}
In standard reinforcement learning (RL) setting, the agent learns to maximize an observed numerical reward signal. However, finding appropriate numerical rewards can often be challenging in practice, and getting rewards right significantly impacts the effectiveness of RL algorithms \citep{wirth2017survey}. To address this challenge, preference-based RL (PbRL) with human feedback has emerged as a promising alternative \citep{christiano2017deep}. In PbRL, the agent does not receive a numerical reward signal, but rather feedback from a human expert in the form of \emph{preferences} for a state-action trajectory in given pairs of trajectories. PbRL has gained considerable attention across multiple application domains, including games \citep{macglashan2017interactive,christiano2017deep,warnell2018deep}, large language models \citep{ziegler2019fine,stiennon2020learning,wu2021recursively,nakano2021webgpt,ouyang2022training,glaese2022improving,bai2022training,ramamurthy2022reinforcement,liu2023languages}, and robot learning \citep{brown2019extrapolating,shin2023benchmarks}.

In this work, we focus on the problem of offline PbRL, where the learning process relies exclusively on pre-collected offline data without active interaction with the environment. Offline RL has gained significant attention in various applications where conducting real-time online experiments may be costly. In the context of PbRL, an offline setting is particularly relevant due to the high cost and latency associated with obtaining human feedback. One of the key challenges in offline RL is the limited coverage of available offline data. Since coverage of the entire state-action space is rarely feasible in practice \citep{chen2019information}, recent empirical and theoretical approaches to offline RL leverage pessimism so as to rely only on the coverage of one comparator policy (possibly the optimal one), \ie, the so-called partial coverage condition \citep{Yu2020,Kidambi2020,RashidinejadParia2021BORL,li2022pessimism,shi2022pessimistic,yin2021towards,xie2021bellman,uehara2021pessimistic,zhan2022offline}. 
In the context of PbRL, it is also crucial to develop algorithms that work under the partial coverage condition. \looseness=-1

Despite its significance, there are very few algorithms specifically designed for offline PbRL with strong statistical guarantees. In this work, we provide such algorithms and guarantees when preferences depend on unknown reward functions over trajectories. Notably, we consider general reward functions that can be defined over the whole trajectory rather than just state-action pairs. This is consistent with many practical settings in natural language processing. For instance, all benchmarks presented in RL4LM \citep{ramamurthy2022reinforcement} use metrics defined over the entire trajectories. Our main contributions can be summarized as follows:
\begin{itemize}[leftmargin=*,topsep=0pt,itemsep=0.5ex,partopsep=0ex,parsep=0ex]
\item We propose a simple algorithm with general function approximation that consists of two main steps: (1) estimate the implicit reward using Maximum Likelihood Estimation (MLE) with general function approximation from offline data and (2) solve a distributionally robust planning problem over a confidence set around the MLE.
\item We prove that our algorithm can effectively compete with a target policy as long as the offline data cover the target policy. Our analysis leverages a newly defined concentrability coefficient which is tailored to PbRL. As the concentrability coefficient differs from that in the standard RL setting where state-action-wise rewards are directly observed, we establish lower bounds that highlight the necessity of our partial coverage condition. To the best of our knowledge, this is the first theoretical separation result between standard offline RL and offline PbRL.
\item We extend the algorithm to the setting where the transition kernel is unknown, where we not only construct confidence sets for the reward function but also for the system dynamics. Notably, even though the reward can be trajectory-wise, we only need to estimate the per-step transition dynamics to ensure efficient learning.
\item We further extend our results to the action-based comparison model, where preferences are defined over individual actions instead of entire trajectories based on the advantage function of the optimal policy \citep{ramachandran2007bayesian,zhu2023principled}. In comparison to the case of the trajectory-wise comparison model, we can establish a partial coverage guarantee using a concentrability coefficient on pairs of state-action pairs rather than trajectories. In this scenario, our sample complexity only scales with a bound on the advantage function, which can be much smaller than a bound on per-trajectory rewards as shown in \citet{ross2011reduction,agarwal2019reinforcement}.
\end{itemize}


\section{Related Work}
\label{sec:related}

\textbf{Preference-based Reinforcement Learning.}
The closest work to ours is \cite{zhu2023principled}, which also studies offline PbRL, but their algorithm and analysis are restricted to linear models. Our algorithm and analysis extend to general function approximation. Indeed, general classes such as neural networks are commonly employed in practice \citep{christiano2017deep,abdelkareem2022advances}. In the special case of linear rewards and preferences over trajectories, while our algorithms differ, our guarantees recover theirs. So, our guarantees are more general; see Section~\ref{sec:known}. Moreover, they only consider the setting where the transition kernel is known, while our work can also handle unknown transitions. Finally, in the case of action-based preferences, \citet{zhu2023principled} cannot provide guarantees with partial coverage, even under their restriction to linear models. We demonstrate how to achieve meaningful guarantees under partial coverage and a soft margin (Assumption~\ref{ass:soft}).

\citet{wirth2017survey} provide a survey of PbRL. PbRL has received considerable attention in theoretical RL \citep{yue2012k,novoseller2020dueling,xu2020preference,pacchiano2021dueling,chen2022human} but the focus is largely on online PbRL.
To the best of our knowledge, \citet{zhu2023principled} is the only previous work to provide theoretical guarantees for offline PbRL. 

\paragraph{Offline RL.}  In offline RL, one of the most critical challenges is addressing the issue of insufficient coverage in the offline data. It is well-known that naive methods are unable to learn the optimal policy in such scenarios \citep{rashidinejad2021bridging}. To tackle this problem, numerous algorithms have been proposed with theoretical guarantees \citep{liu2020provably, kumar20conservative, jin2021pessimism, rashidinejad2021bridging, uehara2021pessimistic,li2022settling,shi2022pessimistic,jin2020pessimism,xie2021bellman,zhan2022offline}. The most relevant work is \cite{uehara2021pessimistic}, which focuses on offline model-based RL with general function approximation. However, their methods cannot be directly applied to PbRL since per-step rewards are not observable in our setting. Furthermore, even in the standard RL setting, the construction of confidence intervals differs between our approach and theirs. Another related paper is \citet{cheng2022adversarially}, which considers the general offline pessimistic RL framework in the standard setting and also subtracts a reference term in their algorithm, similar to ours. However, our motivations for such reference terms are quite different from theirs. Additional detailed comparisons are given in Section~\ref{sec:alg} and Remark~\ref{rem:diff}.

%% file: preliminary.tex
\section{Preliminaries}
We first introduce our offline PbRL setting with general function approximation. 

\textbf{Markov decision processes.} We consider an episodic time-inhomogeneous Markov Decision Process (MDP) denoted by $\Mcal$, which consists of a state space $\Scal$, an action space $\Acal$, an initial state distribution $\rhos\in\Delta_{\Scal}$, and a horizon $H\in\NN^{+}$. At each step $h\in[H-1]$, we use $\Pstar_{h}:\Scal\times\Acal\to \Delta_{\Scal}$ to denote the ground truth transitions.  The ground truth reward function for the entire trajectory is denoted by $\rstar:\Tcal\to [0,\rmax]$, where $\Tcal=(\Scal\times\Acal)^H$ represents the set of all possible trajectories. Note that $\rstar$ is a trajectory-wise reward, which is more general than  state-action-wise rewards commonly considered in standard RL, which is the special case where for some $\{\rstar_h\}_{h=1}^H$ we have $\rstar(\tau)=\sum_{h=1}^H \rstar_h(s_h,a_h)$ for a trajectory $\tau=(s_1,a_1,\cdots,s_H,a_H)$.

A history-dependent policy $\pi:=\{\pi_h\}_{h=1}^H$ is characterized by $\pi_h:(\Scal\times\Acal)^{h-1}\times\Scal\to\Delta_{\Acal}$, specifying the probability of selecting actions for the agent at each step $h\in[H]$ based on the entire history. We denote the set of all such history-dependent policies as $\Pihis$. Given a policy $\pi$, we define its expected reward with respect to a general reward function $r$ and initial and transition distributions $P=\{P_h\}_{h=0}^{H-1}$ as $J(\pi;r,P):=\EE_{\tau\sim(\pi,P)}[r(\tau)]$. Here, $\EE_{\tau\sim(\pi,P)}[\cdot]$ represents the expectation over the trajectory distribution when executing the policy $\pi$ under the transition $P$ starting from $P_0$. 
We use $\EE_{\tau\sim\pi}[\cdot]$ or $\EE_{\pi}[\cdot]$ to denote the special case when $P$ is the ground truth distribution $\Pstar:=\{\Pstar_h\}_{h=0}^{H-1}$.

The optimal policy, denoted $\pis$, is the policy that maximizes the expected reward with respect to the true reward $\rstar$ and system dynamics $\Pstar$, \ie, $\pis:=\arg\max_{\pi\in\Pihis}J(\pi;\rstar,\Pstar)$. As the true reward function $\rstar$ is dependent on the entire trajectory, the optimal policy $\pis$ is generally history-dependent. Thus, designing offline PbRL algorithms that can learn history-dependent policies is crucial. \looseness=-1

For any policy $\pi$,  we can define its state-action visitation measure as follows: $
d^{\pi}_h(s,a)=\PP^{\pi,\Pstar}(s_h=s,a_h=a),\forall h \in [H],$ where $\PP^{\pi,\Pstar}(\cdot)$ denotes the distribution of the trajectory when executing policy $\pi$ in $\Pstar$. We will also use $d^{\pi}(\tau)$ to denote $\PP^{\pi,\Pstar}(\tau)$ for the whole trajectory $\tau$. \looseness=-1

A policy is Markovian if at each step it depends solely on the current state. When the reward is state-action-wise and the policy is Markovian, we can define the associated V- and Q-functions as $
V^{\pi}_h(s)=\EE_{\pi}[\sum_{t=h}^H \rstar_t(s_t,a_t) |s_h=s ],\forall h\in[H],
Q^{\pi}_h(s,a)=\EE_{\pi}[\sum_{t=h}^H \rstar_t(s_t,a_t) |s_h=s,a_h=a ],\,\forall h\in[H].$
It is well-known that when the reward is state-action-wise, the optimal policy $\pis$ is both Markovian and deterministic. Furthermore, we have $V^{\pis}_h(s)=\sup_{\pi}V^{\pi}_h(s)$ and $Q^{\pis}_h(s,a)=\sup_{\pi}Q^{\pi}_h(s,a)$ for all $h\in[H]$. For brevity, we will use $\VS$ and $\QS$ to represent the optimal state-value function and Q-function, respectively. The advantage function of the optimal policy, denoted by $\AS$, is defined to be $\AS_h(s,a)=\QS_h(s,a)-\VS_h(s)$ for all $h\in[H],s\in\Scal,A\in\Acal$.

\textbf{Offline Preference-based Reinforcement Learning.} We focus on the problem of offline PbRL in this work. Specifically, in the trajectory-based pairwise comparison setting, we are provided with an offline dataset $\Dcal=\{\tau^{n,0},\tau^{n,1},o^n\}_{n=1}^N$, where $\tau^{n,0}=\{s^{n,0}_{h},a^{n,0}_h\}_{h=1}^H$ and $\tau^{n,1}=\{s^{n,1}_{h},a^{n,1}_h\}_{h=1}^H$ are i.i.d. sampled from the distributions $\mu_{0}$ and $\mu_{1}$, respectively, and $o^n\in\{0,1\}$ indicates preference for $\tau^{n,1}$ over $\tau^{n,2}$. We assume it satisfies the following preference model:
\begin{assumption}[Preference-based model]\label{ass:preference}
Given a pair of trajectories $(\tau^0, \tau^1)$,
 $o\in\{0,1\}$ satisfies
\begin{align*}
 P(o=1\mid \tau_0,\tau_1)=P(\text{$\tau_1$ is preferred over $\tau_0$}\mid \tau_0,\tau_1)=\Phi(r^{\star}(\tau_1)-r^{\star}(\tau_0)).
\end{align*}
where $\Phi:\mathbb{R}\to [0,1]$ is a monotonically increasing link function. 
\end{assumption}
A commonly used link function is the sigmoid function $\sigma(x) = 1/\{1 + \exp(-x)\}$, leading to the Bradley-Terry-Luce (BTL) model \citep{christiano2017deep}. 

The objective of offline PbRL is to learn a high-quality  policy $\hpi \in\Pihis$, \ie, with
$J(\pit;\rstar,\Pstar)-J(\hpi;\rstar,\Pstar)\leq\epsilon$ where $\pit$ is a target policy we want to compete with (potentially $\pis$).

\textbf{General function approximation.} 
In our paper, we estimate the reward $\rstar$ with general function approximation. We introduce a function class $\Gcal_r$, such as linear functions or neural networks, to approximate the true reward. For each $r \in \Gcal_r$ and trajectory pair $(\tau^0, \tau^1)$, we denote the induced preference model with respect to $r$ as $P_r(o|\tau^0,\tau^1)$, defined as
\begin{align}
\label{eq:Pr}
    P_r(o=1\mid \tau^0,\tau^1):=\Phi(r(\tau^1)-r(\tau^0)). 
\end{align} 
We use bracketing numbers to measure the complexity of $\{P_r:r\in\Gcal_r\}$.
\begin{definition}[$\eps$-bracketing number of preferences]
We say $(g^1,g^2)$ is an $\eps$-bracket if $g^1(\cdot\mid\tau^0,\tau^1)\leq g^2(\cdot\mid\tau^0,\tau^1)$ and $\Vert g^1(\cdot\mid\tau^0,\tau^1)-g^2(\cdot\mid\tau^0,\tau^1)\Vert_1\leq\eps$ for all trajectory-pairs $(\tau^0,\tau^1)$. The $\eps$-bracketing number of $\Gcal_r$, denoted by  $\Ncal_{\Gcal_r}(\eps)$, is the minimal number of $\eps$-brackets $(g^{n,1},g^{n,2})_{n=1}^N$ needed so that for any $r\in\Gcal_r$ there is a bracket $i\in[N]$ containing it, meaning $g^{i,1}(\cdot|\tau^0,\tau^1)\leq P_r(\cdot| \tau^0,\tau^1)\leq g^{i,2}(\cdot|\tau^0,\tau^1)$ for all trajectory-pairs $(\tau^0,\tau^1)$. 
\end{definition}

The $\epsilon$-bracket number is widely used in statistics \citep{geer2000empirical} to study MLE and related M-estimates. Particularly, in our setting the bracket number of reward classes will be of the same order as the covering number, another common complexity measure in statistics \citep{wainwright2019high}, for $P_r(\cdot|\tau^0,\tau^1)$ has only two dimensions. One example for which we can bound the $\epsilon$-bracket number is linear rewards under the BTL model \citep{pacchiano2021dueling, zhu2023principled}. 


\begin{proposition}
\label{prop:bracket-linear}
Suppose $\Vert\phi(\tau)\Vert_2\leq R$ $\forall \tau\in\Tcal$, $\Gcal_r\subseteq\{\tau\mapsto\langle \phi(\tau),\theta\rangle:\Vert\theta\Vert_2\leq B\}$ for some featurization $\phi:\Tcal\to\RR^d$ and $B>0$, and the link function is $\Phi(\cdot)=\sigma(\cdot)$. Then for any $\eps\leq1$,
$
\log\Ncal_{\Gcal_r}(\eps)\leq\Ocal(d\log\frac{BR}{\eps}).
$
\end{proposition}
The proof is deferred to Appendix~\ref{proof:prop-bracket-linear}.
To handle unknown transitions, we use function classes $\{\Gcal_{P_h}\}_{h=0}^{H-1}$
to approximate the transition probabilities $\{\Pstar_h\}_{0=1}^{H-1}$. 
Similarly, we use $\Ncal_{\Gcal_{P_h}}(\eps)$
to denote the $\eps$-bracket number of $\Gcal_{P_h}$, which is defined as follows:
\begin{definition}[$\eps$-bracket number of transition probability classes]
Suppose $f^1,f^2$ is a function with $f^1(\cdot|s,a),f^2(\cdot|s,a)\in\RR^{|\Scal|}$ for all $(s,a)\in\Scal\times\Acal$. Then we say $(f^1,f^2)$ is a $\eps$-bracket if $f^1(\cdot|s,a)\leq f^2(\cdot|s,a)$ and $\Vert f^1(\cdot|s,a)-f^2(\cdot|s,a)\Vert_1\leq\eps$ for all $(s,a)$. The $\eps$-bracket number of a transition probability class $\Gcal_{P_h}$ where $h\in[H-1]$ is the minimum integer $N$ satisfying that there exist $N$ $\eps$-brackets $(f^{n,1},f^{n,2})_{n=1}^N$ such that for any function $P_h\in\Gcal_{P_h}$ there is a bracket $(f^{i,1},f^{i,2})$ where $i\in[N]$ containing it, i.e., $f^{i,1}(\cdot|s,a)\leq P_h(\cdot|s,a)\leq f^{i,2}(\cdot|s,a)$ for all $(s,a)$. 
\end{definition}
\begin{definition}[$\eps$-bracket number of initial state distribution classes]
Suppose $f^1,f^2\in\RR^{|\Scal|}$. Then we say $(f^1,f^2)$ is a $\eps$-bracket if $f^1\leq f^2$ and $\Vert f^1-f^2\Vert_1\leq\eps$. The $\eps$-bracket number of a initial state distribution class $\Gcal_{P_0}$ is the minimum integer $N$ satisfying that there exist $N$ $\eps$-brackets $(f^{n,1},f^{n,2})_{n=1}^N$ such that for any $\rho\in\Gcal_{P_0}$ there is a bracket $(f^{i,1},f^{i,2})$ where $i\in[N]$ containing it, i.e., $f^{i,1}\leq \rho\leq f^{i,2}$. 
\end{definition}
When the transition probability possesses a low-dimension embedding, we can also bound the $\eps$-bracket number of the function class efficiently.

%% file: known.tex
\section{Trajectory-Based Pairwise-Comparison with Known Transition}
\label{sec:known}
In this section, we present our algorithm and analyze the sample complexity for the trajectory-based pairwise-comparison setting when the ground truth transition $\Pstar$ is known. In Sections~\ref{sec:unknown} and \ref{sec:action}, we will further explore the unknown transition setting and the action-based comparison setting.
\subsection{Algorithm}
\label{sec:alg}
Our proposed algorithm, \mainalg described in Algorithm~\ref{alg:main},
consists of the following two steps.

\textbf{Confidence set construction via MLE (Lines~\ref{line:MLE}--\ref{line:MLE2}).} We construct a confidence set for the ground truth reward from the implicit preference feedback. We achieve this by selecting reward models that nearly maximize the log-likelihood of observed data up to a slackness parameter $\zeta$. We will show that the result, $\Rcal(\Dcal)$, approximates the following confidence set:
\begin{align*}
     \Rcal'(\Dcal):=\{r\in \Gcal_r:\mathbb{E}_{\tau_0\sim \mu_0,\tau_1 \sim \mu_1}[|\{r(\tau_1)-r(\tau_0)\}-\{r^{*}(\tau_1)-r^{*}(\tau_0)\}|^2]\leq \xi\}
\end{align*}
for a certain $\xi$. Here the distance between $r$ and $r^{\star}$ is measured using the total variation distance (\ie, $\ell_1$ norm) of $r(\tau_1)-r(\tau_0)$ and  $r^*(\tau_1)-r^*(\tau_0)$ over the offline data. 

\textbf{Distributionally robust policy optimization (Line~\ref{line:dist}).} After constructing the confidence set, we search for the policy that maximizes the policy value under the least favorable reward model, the $r \in \Rcal(\Dcal)$ minimizing the policy value $J(\pi;r,P^*)$ minus $\EE_{\tau\sim\muref}[r(\tau)]$, where $\muref$ is an arbitrary known reference trajectory distribution. 
It is generally recommended to set $\muref$ to $\mu_1$, as we will explain later, possibly a sample-average approximation thereof based on $\{\tau^{1,1},\dots,\tau^{N,1}\}$.
By selecting the least favorable reward model instead of the MLE solution $\hr$, we penalize policies that are not well-covered by the offline data. 
The need for a reference policy arises because the approximated confidence set measures the uncertainty for reward difference between two trajectories ($r(\tau_1)-r(\tau_0)$), but it cannot measure the uncertainty of the reward of a single trajectory.

In the following, we compare our algorithm to existing works. \cite{zhu2023principled} consider a pessimistic offline RL algorithm for PbRL specialized to the linear reward class setting, while our \mainalg can handle general function approximation. Specifically, they construct the confidence set using the feature-covariance-rotated $\ell_2$-ball around the MLE $\hat\theta$, where $\hr(\tau)=\langle \phi(\tau),\hat\theta\rangle$. In contrast, our confidence set is obtained directly from the log-likelihood objective and is generic. \citet{uehara2021pessimistic} proposes a model-based pessimistic offline RL algorithm when we have access to rewards. The confidence set construction correspondingly differs significantly. \citet{cheng2022adversarially} considers a general offline pessimistic RL framework. In their policy optimization step, they also subtract the value of a reference policy. This similarity is superficial, however, as the motivations are different. We subtract the value because we can only measure the difference between rewards of any two trajectories, while their motivation is to obtain a certain robustness result (their proposition 3).

\begin{remark}[Computational Efficiency]
Line 4 in \mainalg is computationally hard in general. Nevertheless, by leveraging Lagrangian formulation, we can use Lagrangian multiplier to convert the constraint $r\in\Rcal(\Dcal)$ into a regularization term of the objective function and have a feasible version of our algorithm in practice. See more details in Appendix~\ref{sec:discussion}.
\end{remark}

\begin{algorithm}[t!]
	\caption{\textbf{\mainalg}: oFfline ReinforcemEnt lEarning with HumAN feeDback}
	\label{alg:main}
	\begin{algorithmic}[1]
		\State \textbf{Input}: offline datset $\Dcal$, slackness parameter $\zeta$, reference distribution $\muref$, true transition $\Pstar$ 
		\State 	\textbf{MLE}: compute  $\hr = \argmax_{r \in \Gcal_r} \sum_{n=1}^N \log P_r(o=o^{n}\mid\tau^{n,1},\tau^{n,0})$\label{line:MLE}
		\State  \textbf{Confidence set construction}: construct 
        $$\textstyle\Rcal(\Dcal)=\big\{r\in \Gcal_r: \sum_{n=1}^N \log P_r(o=o^{n}\mid\tau^{n,0},\tau^{n,1})\geq \sum_{n=1}^N \log P_{\hr}(o=o^{n}\mid\tau^{n,0},\tau^{n,1})-\zeta \big\}$$  \label{line:MLE2}
		\State \textbf{Distributionally robust planning}: return
$$\textstyle\hpi = \argmax_{\pi\in\Pihis }\min_{r \in \Rcal(\Dcal)} \left( J(\pi; r,\Pstar)-\EE_{\tau\sim\muref}[r(\tau)]\right)$$   \label{line:dist}
	\end{algorithmic}
\end{algorithm}

\subsection{Analysis}
To analyze the sample complexity of \mainalg, we first quantify the discrepancy between the offline data $\Dcal$ and the distribution induced by the target policy $\pit$.
\begin{definition}[concentrability coefficient for preference-based feedback]
\label{def:concentrability}
The concentrability coefficient w.r.t. a reward class $\Gcal_r$, a target policy $\pit$, and a reference policy $\muref$ is defined as 
\begin{align*}
C_r(\Gcal_r,\pit,\muref):=\max\Biggl\{0,~\sup_{r\in\Gcal_r}\frac{\EE_{\tau^0 \sim \pit,\tau^1 \sim \muref}[\rstar(\tau^0)-\rstar(\tau^1)-r(\tau^0)+r(\tau^1)]}{\sqrt{\EE_{\tau^0 \sim \mu_0,\tau^1 \sim \mu_1}\big[|\rstar(\tau^0)-\rstar(\tau^1)-r(\tau^0)+r(\tau^1)|^2\big]}}\Biggr\}.
\end{align*} 
\end{definition} 
Note, when we choose $\muref=\mu_1$, by Jensen's inequality, the value of $C_r(\Gcal_r,\pit,\mu_1)$ can always be upper bounded by the per-trajectory concentration coefficient: $C_{r}(\Gcal_r, \pit,\mu_1)\leq \sqrt{\Ctr}$ for any $\Gcal_r$, where $\textstyle \Ctr:=\max_{\tau\in\Tcal}\frac{d^{\pit}(\tau)}{\mu_0(\tau)}.$ Moreover, while $C_r(\Gcal_r,\pit,\mu_1)$ becomes $\sqrt{\Ctr}$ in the worst case (\eg, when $\Gcal_r$ is the set of all functions mapping from $\Tcal$ to $\mathbb{R}$), it can generally be much smaller. For example, when using linear models, it is a relative condition number, as explained later. Finally, when $\muref=d^{\pit}$, our coefficient becomes $0$. This implies that $C_r(\Gcal_r,\pit,\mu_1)$ could be small when $\pit$ and $\muref$ are close. While the concept of concentrability coefficient has been used in offline RL with explicit reward feedback \citep{chen19information,song2022hybrid}, this property is unique when the feedback is in the form of preferences.

In our following PAC analysis, we further assume the reward class $\Gcal_r$ is realizable and bounded. 
\begin{assumption}[Realizability]
\label{ass:realize}
We have $\rstar\in\Gcal_r$.
\end{assumption}

\begin{assumption}[Boundedness]
\label{ass:bound}
We have $0\leq r(\tau)\leq\rmax$ for all $r\in\Gcal_r$ and $\tau\in\Tcal$.
\end{assumption}

\begin{theorem}
\label{thm:main}
For any $\delta\in(0,1]$, let $\zeta=\cmle\log(\Ncal_{\Gcal_r}(1/N)/\delta)$ where $\cmle>0$ is a universal constant, then under Assumption~\ref{ass:preference},\ref{ass:realize} and \ref{ass:bound}, with probability $1-\delta$, we have
\begin{align}\label{eq:main}
J(\pit;\rstar,\Pstar)-J(\hpi; \rstar, \Pstar)\leq\sqrt{\frac{cC^2_{r}(\Gcal_r, \pit,\muref)\kappa^2\log(\Ncal_{\Gcal_r}(1/N)/\delta)}{N}},
\end{align}
where $c>0$ is a universal constant and $\kappa=(\inf_{x\in[-\rmax,\rmax]}\Phi'(x))^{-1}$.
\end{theorem}

Theorem~\ref{thm:main} indicates that \mainalg can learn an $\eps$-optimal policy compared to $\pit$ with a sample complexity of $$N=\TO\bigg(\frac{C^2_{r}(\Gcal_r,\pit,\muref)\kappa^2\log(\Ncal_{\Gcal_r}(1/N)/\delta)}{\eps^2}\bigg).$$ Next we provide a detailed explanation of this sample complexity. Firstly, $C_r(\Gcal_r,\pit,\muref)$ represents the extent to which the dataset $\Dcal$ covers the target policy $\pit$. In our theorem, to obtain a non-vacuous PAC guarantee, we only require the dataset $\Dcal$ to cover the target policy $\pit$ (i.e., $C_r(\Gcal_r,\pit,\muref)<\infty$). The distributionally robust optimization step plays a crucial role in obtaining this guarantee under partial coverage. In particular, invoking the abovementioned third property of $C_r(\Gcal_r,\pit,\muref)$, when setting $\pit=\muref$, \pref{eq:main} is reduced to 
\begin{align}
J(\muref;\rstar,\Pstar)\leq J(\hpi; \rstar, \Pstar)
\end{align}
This encourages us to choose $\muref=\mu_1$ (or $\mu_0)$ as it will allow us to ensure our performance is at least larger than the performance associated with the offline data.

Secondly, $\log(\Ncal_{\Gcal_r}(1/N))$ measures the complexity of the function class $\Gcal_r$. For example, when using linear models, it takes $\tilde O(d)$. We refer the reader to \citet{geer2000empirical} for bracketing number computations for more general classes. 
Thirdly, $\kappa$ represents the non-linearity of the link function $\Phi$, which determines the difficulty of estimating the reward from human preferences. This dependence on $\kappa$ is present in the existing literature of PbRL, both in online settings \citep{pacchiano2021dueling,chen2022human} and offline settings \citep{zhu2023principled}.

\paragraph{Comparison with \cite{zhu2023principled}.} \cite{zhu2023principled} is the most related work to our paper. They consider the linear reward setting under BTL model and can achieve the following sample complexity:
\begin{align*}
N=\Ocal\bigg(\frac{C^2_{\mathrm{lin}}\exp(4BR)d\log(1/\delta)}{\eps^2}\bigg),
\end{align*}
where $R$ and $B$ are the norm bounds on the feature vectors $\phi$ and parameter $\theta$ (defined in Proposition~\ref{prop:bracket-linear}).The concentrability coefficient $C_{\mathrm{lin}}$ is defined as
\begin{align*}
C_{\mathrm{lin}}:=\Vert\EE_{\tau^0\sim\pit,\tau^1\sim\muref}[\phi(\tau^0)-\phi(\tau^1)]\Vert_{\Sigma_{\Dcal}^{-1}},
\end{align*}
and $\Sigma_{\Dcal}$ is the empirical covariance matrix of the dataset $\frac{1}{N}\sum_{n=1}^N(\phi(\tau^{n,0})-\phi(\tau^{n,1}))(\phi(\tau^{n,0})-\phi(\tau^{n,1}))^{\top}$.

Note that all the analysis and proofs in this paper still hold when we define the concentrability coefficient as
\begin{align*}
C'_r(\Gcal_r,\pit,\muref):=\max\bigg\{0,\sup_{r\in\Gcal_r}\frac{\EE_{\tau^0 \sim \pit,\tau^1 \sim \muref}[\rstar(\tau^0)-\rstar(\tau^1)-r(\tau^0)+r(\tau^1)]}{\sqrt{\frac{1}{N}\sum_{n=1}^N|\rstar(\tau^{n,0})-\rstar(\tau^{n,1})-r(\tau^{n,0})+r(\tau^{n,1})|^2}}\bigg\}.
\end{align*}
Then when specializing the result in Theorem~\ref{thm:main} to the linear reward setting under BTL model with this version of concentrability coefficient, the sample complexity is
\begin{align*}
N=\TO\bigg(\frac{(C'_{r}(\Gcal_r,\pit,\muref))^2\exp(2\rmax)d\log(BR/\delta)}{\eps^2}\bigg).
\end{align*}
We know that $BR\geq\rmax$. In addition, note that in this case, we have $C_{\mathrm{lin}}\geq0$ and for any $r\in\Gcal_r$,
\begin{align*}
&\big|\EE_{\tau^0 \sim \pit,\tau^1 \sim \muref}[\rstar(\tau^0)-\rstar(\tau^1)-r(\tau^0)+r(\tau^1)]\big|\\
&\qquad=\big|\langle\EE_{\tau^0\sim\pit,\tau^1\sim\muref}[\phi(\tau^0)-\phi(\tau^1)],\theta^{\star}-\theta\rangle\big|\\
&\qquad\leq\Vert\EE_{\tau^0\sim\pit,\tau^1\sim\muref}[\phi(\tau^0)-\phi(\tau^1)]\Vert_{\Sigma_{\Dcal}^{-1}}\cdot\Vert\theta^{\star}-\theta\Vert_{\Sigma_{\Dcal}}\\
&\qquad=\Vert\EE_{\tau^0\sim\pit,\tau^1\sim\muref}[\phi(\tau^0)-\phi(\tau^1)]\Vert_{\Sigma_{\Dcal}^{-1}}\cdot\sqrt{\frac{1}{N}\sum_{n=1}^N|\rstar(\tau^{n,0})-\rstar(\tau^{n,1})-r(\tau^{n,0})+r(\tau^{n,1})|^2},
\end{align*}
where we suppose $\rstar(\tau)=\langle\phi(\tau),\theta^{\star}\rangle$ and $r(\tau)=\langle\phi(\tau),\theta\rangle$. Therefore we have
\begin{align*}
C'_{r}(\Gcal_r,\pit,\muref)\leq C_{\mathrm{lin}}.
\end{align*}
This implies that Theorem~\ref{thm:main} can recover the sample complexity for linear reward setting under BTL model in \cite{zhu2023principled} with only some additional log factors.

\begin{remark}  In practice, to compute $\mathbb{E}_{\tau \sim \mu_1}[r(\tau)]$ in Line~\ref{line:MLE2}, we can use the sample average, with an additional cost of $\sqrt{\log(1/\delta)/N}$ in the suboptimality bound in Eq.~\pref{eq:main}.
\end{remark}

\subsection{Discussion of the Concentrability Coefficient}
In the worst-case scenario (\ie, $\Gcal_r$ is the set of all functions mapping from $\Tcal$ to $\mathbb{R}$), the value of $C_r(\Gcal_r,\pit,\mu_1)$ is reduced to to the per-trajectory concentrability coefficient $\Ctr$. The per-trajectory concentrability coefficient is generally larger than the per-step concentrability coefficient $\Cst$ commonly used in the general offline RL literature. Specifically, $\Cst$ is defined as $$\Cst:=\max_{s,a,h} d^{\pit}_h(s,a)/\mu_{0,h}(s,a),$$ where $\mu_{0,h}(s,a)$ represents the marginal distribution at step $h$. In this section, we show the dependence on the per-trajectory concentrability coefficient is necessary for our offline PbRL context. This is intuitively because our PbRL setting involves reward functions defined over trajectories, reflecting the fact that human feedback is also trajectory-based.

In the next proposition, we first show that the per-trajectory concentrability coefficient $\Ctr$ can be exponentially larger than the per-step concentrability coefficient $\Cst$.

\begin{proposition}
\label{prop:tr}
For any $S\geq1, A\geq2,H\geq1, C\geq1$, there exists an MDP $\Mcal$ with horizon $H$, a policy $\pit$ and a data distribution $\mu_0$ such that $|\Scal|=S,|\Acal|=A$ and $\Cst=C$ while $\Ctr=C^H$.
\end{proposition}
Proposition~\ref{prop:tr} indicates that $\Ctr$ can be significantly larger than $\Cst$.  A natural question arises as to whether we can obtain suboptimality guarantees using $\Cst$. Unfortunately, the following lower bounds reveal that per-step concentrability is not sufficient to guarantee efficient learning in trajectory-based PbRL setting, even when the reward function is defined over state-action pairs:
\begin{theorem}
\label{thm:lower}
Set $\pit=\pis$. Then, for any $C>1$ and $H\geq2$, there exists a dataset distribution $\mu_1$ such that we have
\begin{align*}
\inf_{\hpi}\sup_{(\Mcal,\mu_0)\in\overline{\Theta}_{\mathrm{st}}(C)}\EE_{\Dcal}[J(\pis;\rstar,\Pstar)-J(\hpi;\rstar,\Pstar)]\gtrsim\min\bigg\{C-1,1\bigg\}, 
\end{align*}
where $\hpi$ is any mesurable function of the data $\Dcal$ (and knows the information of $\mu_1$). 
$\overline{\Theta}_{\mathrm{st}}(C)$ is the set of all MDPs with per-step reward, offline distribution $(\Mcal,\mu_0)$ such that $\Cst\leq C$. Note $\EE_{\Dcal}$ is taken with respect to the randomness in $\Dcal$.  
\end{theorem}


Theorem~\ref{thm:lower} indicates that learning in our setting is intrinsically hard due to trajectory-based feedback, even if we restrict the reward function class. In addition, we can show that scaling with $\Ctr$ is necessary in our setting:
\begin{theorem}
\label{thm:lower2}
Set $\pit=\pis$. Then for any $C>1$ and $H\geq1$, there exists a dataset distribution $\mu_1$ such that we have
\begin{align*}
\inf_{\hpi}\sup_{(\Mcal,\mu_0)\in\Theta_{\mathrm{tr}}(C)}\EE_{\Dcal}[J(\pis;\rstar,\Pstar)-J(\hpi;\rstar,\Pstar)]\gtrsim\min\bigg\{C-1,\sqrt{\frac{C-1}{N}}\bigg\},
\end{align*}
where $\hpi$ is any mesurable function of the data $\Dcal$ (and knows the information of $\mu_1$). $\Theta_{\mathrm{tr}}$ is the set of all MDP, offline distribution $(\Mcal,\mu_0)$ such that $\Ctr\leq C$. Note $\EE_{\Dcal}$ is taken with respect to the randomness in $\Dcal$. 
\end{theorem}

Note that when $\mu_1$ is known, we can set $\muref=\mu_1$ in Algorithm~\ref{alg:main} and then $C_r(\Gcal_r,\pit,\mu_1)\leq\sqrt{\Ctr}$, which implies the sample complexity in Theorem~\ref{thm:main} indeed nearly matches this lower bound with respect to $\Ctr$ and $N$ when $N$ is sufficiently large.  

In summary, Theorem~\ref{thm:lower} and Theorem~\ref{thm:lower2} imply that the scaling with the per-trajectory concentrability coefficient is essential in the trajectory-based pairwise-comparison setting, and it cannot be relaxed to the per-step concentrability without additional assumptions, such as on the reward structure. To the best of our knowledge, this is the first theoretical result indicating that trajectory-wise feedback is \textbf{intrinsically harder} to learn than step-wise feedback in offline PbRL.

%% file: unknown.tex
\section{Trajectory-Based Comparison with Unknown Transition}
\label{sec:unknown}
We extend the setting presented in Section~\ref{sec:known} to the scenario where the transition function $\Pstar$ is unknown. The algorithm is described in Algorithm~\ref{alg:un}. Compared to Algorithm~\ref{alg:main}, we simply added a similar step to handle unknown transitions. 
Hereafter, we use the convention $P_{0}(\cdot\mid s,a):=P_{0}(\cdot)$.

\begin{algorithm}[!t]
	\caption{\textbf{\unalg}}
	\label{alg:un}
	\begin{algorithmic}
		\State \textbf{Input}: offline dataset $\Dcal$, slackness parameter $\zeta,\zeta_{P_h}$, reference distribution $\muref$
  
	\State 	\textbf{MLE for reward}: compute $\textstyle \hr = \argmax_{r \in \Gcal_r} \sum_{n=1}^N \log P_r(o=o^{n}|\tau^{n,1},\tau^{n,0})$
\State 	\textbf{MLE for transition}: 
compute $\hP_h = \argmax_{P_h \in \Gcal_{P_h}} \sum_{n=1}^N\sum_{i=0}^1 \log P_h(s^{n,i}_{h+1}|s^{n,i}_h,a^{n,i}_h)$
		\State  \textbf{Confidence set construction}: for $0\leq h\leq H-1$, construct
        $$\textstyle \Rcal(\Dcal)=\bigg\{r\in \Gcal_r: \sum_{n=1}^N \log P_r(o=o^{n}|\tau^{n,0},\tau^{n,1})\geq \sum_{n=1}^N \log P_{\hr}(o=o^{n}|\tau^{n,0},\tau^{n,1})-\zeta \bigg\}.$$
        $$\Pcal_h(\Dcal)=\bigg\{P_h\in \Gcal_{P_h}: \sum_{n=1}^N\sum_{i=0}^1 \log P_h(s^{n,i}_{h+1}|s^{n,i}_h,a^{n,i}_h)\geq \sum_{n=1}^N\sum_{i=0}^1 \log \hP_h(s^{n,i}_{h+1}|s^{n,i}_h,a^{n,i}_h)-\zeta_{P_h} \bigg\},$$
        
		\State \textbf{Distributionally robust plnanning}: return 
$$\textstyle \hpi = \argmax_{\pi\in\Pihis }\min_{r \in \Rcal(\Dcal),P_h\in\Pcal_h(\Dcal)} J\Big(\pi; r,\{P_h\}_{h=0}^{H-1})\Big)-\EE_{\tau\sim\muref}[r(\tau)].$$

	\end{algorithmic}
\end{algorithm}

Our sample complexity will depend on the following additional concentration coefficient:
\begin{definition}[Concentrability coefficient for the transition]
\label{def:concentrability2}
The concentrability coefficient w.r.t. transition classes $\{\Gcal_{P_h}\}$ and a target policy $\pit$ is defined as 
\begin{align*}
C_P(\{\Gcal_{P_h}\},\pit):=\max_{h:0\leq h\leq H-1}\sup_{P_h \in \Gcal_{P_h}}\frac{ \EE_{(s,a)\sim d^{\pit}_h}[\Vert P_h(\cdot \mid s,a) -P^{\star}_h(\cdot \mid s,a)\Vert_1  ] }{\sqrt{\EE_{(s,a)\sim (\mu_{0,h}/2 + \mu_{1,h}/2) }[\Vert P_h(\cdot \mid s,a) -P^{\star}_h(\cdot \mid s,a)  \Vert^2_1]}}.
\end{align*}
\end{definition}
Note this is always upper-bounded by the density-ratio-based concentrability coefficient, $C_P(\{\Gcal_{P_h}\},\pit)\leq\sup_{(s,a,h)\in\Scal\times\Acal\times[H]}\frac{d^{\pit}_h(s,a)}{\mu_{0,h}(s,a)/2+\mu_{1,h}(s,a)/2}.$

We also assume the transition classes $\{\Gcal_{P_h}\}_{h=0}^{H-1}$ are realizable:
\begin{assumption}[Realizability]
\label{ass:realize2}
Suppose that we have $\Pstar_h\in\Gcal_{P_h}$ for all $h$ where $0\leq h\leq H-1$. In addition, any choice $P_h\in\Gcal_{P_h}$ for $0\leq h\leq H-1$ are valid transition distributions. 
\end{assumption}

Then with the above assumptions,  we have the following theorem to characterize the sample complexity when the transition is unknown:
\begin{theorem}
\label{thm:unknown}
For any $\delta\in(0,1]$, let $\zeta=\cmle\log(\Ncal_{\Gcal_r}(1/N)/\delta)$,$\zeta_{P_h}=c_P\log(H\Ncal_{\Gcal_{P_h}}(1/N)/\delta)$ 
where $\cmle,c_P>0$ are universal constants, then under Assumption~\ref{ass:preference},\ref{ass:realize},\ref{ass:bound} and \ref{ass:realize2}, we have
\begin{align*}
&J(\pit;\rstar,\Pstar)-J(\hpi; \rstar, \Pstar)\\
&\leq\sqrt{\frac{cC^2_{r}(\Gcal_r,\pit,\muref)\kappa^2\log(\Ncal_{\Gcal_r}(1/N)/\delta)}{N}}+H\rmax\sqrt{\frac{cC^2_P(\{\Gcal_{P_h}\},\pit)\log(H\Ncal_{P}(1/N)/\delta)}{N}},
\end{align*}
where $c>0$ and $\kappa$ are the same as Theorem~\ref{thm:main} and $\Ncal_P:= \max_{0\leq h\leq H-1}\Ncal_{\Gcal_{P_h}}$.
\end{theorem}
Compared to Theorem~\ref{thm:main}, we introduce an additional term in our guarantee to account for the unknown transitions. Once again, our result demonstrates that the learned policy can achieve performance comparable to any target policy $\pit$ covered by the offline data, i.e., $C_{r}(\Gcal_r,\pit,\muref)<\infty,C_P(\{\Gcal_{P_h}\},\pit)<\infty$.

\begin{remark}[Comparison to \citet{uehara2021pessimistic} ]\label{rem:diff}
Like us, \citet{uehara2021pessimistic} proposed a model-based RL algorithm that works under partial coverage, but in the standard RL setting and with a known state-action-wise reward function. In addition to the difference in settings, which is the primary difference, our approach moreover differs from their approach because while they construct confidence intervals by defining a confidence ball around the MLE solution based on the total variation distance, we use the Kullback-Leibler (KL) distance. This may be preferable as computing the KL distance is generally easier than the total variation distance as it arises directly from the MLE objective, as practically done in \citet{rigter2022rambo}. 
\end{remark}

%% file: action.tex
\section{Action-Based Comparison }
\label{sec:action}
 Next, we turn our attention to the action-based comparison setting \citep{ramachandran2007bayesian,zhu2023principled}, where human evaluators provide preferences between pairs of actions instead of pairs of trajectories. In this section, we assume that the reward function $\rstar$ is state-action-wise:  $\rstar(\tau) = \sum_{h=1}^H \rstar_h(s_h,a_h)$ for $\tau = (s_1,a_1,\cdots,s_H,a_H)$. And, we consider a preference model based on $\QS$.

 \textbf{Setting.} We have datasets $\Dcal=\{\Dcal_h\}_{h=1}^H$ with $\Dcal_h=\{(s_h^{n},a_{h}^{n,0},a_{h}^{n,1},o_h^n)\}_{n=1}^N$ for each $h\in[H]$, where each sample is drawn i.i.d.  from the distribution $s_h^n\sim\mu_h, a_h^{n,0}\sim\mu_{0,h}(\cdot\mid s_h^n), a_h^{n,1}\sim\mu_{1,h}(\cdot\mid s_h^n)$ and $o^n_h\in\{0,1\}$ indicates preference for $a_h^{1,n}$ over $a_h^{0,n}$ in the state $s_h^n$.
 We assume it satisfies the following preference model:
 \begin{assumption}[Action-based comparison model] \label{ass:preference_action}
 Given a pair of actions $a^0_h,a^1_h$ and state $s_h$, $o\in\{0,1\}$ satisfies
\begin{align*}
    P(o_h=1\mid s_h,a_{h}^{0},a_{h}^{1})= \Phi(\QS_h(s_h,a_h^{1})-\QS_h(s_h,a_h^{0})). 
\end{align*}
 \end{assumption}
Here, the aforementioned distribution can be equivalently expressed as $ P(o_h^n=1\mid s_h^n,a_{h}^{n,0},a_{h}^{n,1})= \Phi(\AS_h(s_h^n,a_h^{n,1})-\AS_h(s_h^n,a_h^{n,0}))$, where $\AS$ denotes the optimal advantage function. Consequently, we introduce general function classes $\Gcal_{A_h}$ to estimate the optimal advantage function $\AS_h$. In addition, for each $A_h\in\Gcal_{A_h}$ and $(s,a^0,a^1)\in\Scal\times\Acal\times\Acal$, we use $P_{A_h}(\cdot\mid s,a^0,a^1)$ to represent the human preference model with respect to $A_h$, defined as $ P_{A_h}(o=1\mid s,a^0,a^1):=\Phi(A_h(s,a^1)-A_h(s,a^0)). $

We again use the $\eps$-bracket number of such advantage function classes to quantify their complexity, denoted as $\Ncal_{\Gcal_{A_h}}$:
\begin{definition}[$\eps$-bracket number of advantage function classes]
Suppose $g^1,g^2$ is a function with $g^1(\cdot|s,a^0,a^1),g^2(\cdot|s,a^0,a^1)\in\RR^2$ for all $(s,a^0,a^1)\in\Scal\times\Acal\times\Acal$. Then we say $(g^1,g^2)$ is a $\eps$-bracket if $g^1(\cdot|s,a^0,a^1)\leq g^2(\cdot|s,a^0,a^1)$ and $\Vert g^1(\cdot|s,a^0,a^1)-g^2(\cdot|s,a^0,a^1)\Vert_1\leq\eps$ for all $(s,a^0,a^1)\in\Scal\times\Acal\times\Acal$. The $\eps$-bracket number of a reward class $\Gcal_{A_h}$ where $h\in[H]$ is the minimum integer $N$ satisfying that there exist $N$ $\eps$-brackets $(g^{n,1},g^{n,2})_{n=1}^N$ such that for any function $A_h\in\Gcal_{A_h}$ there is a bracket $(g^{i,1},g^{i,2})$ where $i\in[N]$ containing it, i.e., $g^{i,1}(\cdot|s,a^0,a^1)\leq P_{A_h}(\cdot| s,a^0,a^1)\leq g^{i,2}(\cdot|s,a^0,a^1)$ for all $(s,a^0,a^1)\in\Scal\times\Acal\times\Acal$. 
\end{definition} 
Similarly, when the advantage function possesses a low-dimension embedding, we can also bound the $\eps$-bracket number efficiently.

\subsection{Algorithm}
Our algorithm comprises two steps. In the first step (Line~\ref{line:MLE_Q}), our objective is to estimate the optimal advantage function using MLE. In the second step (Line~\ref{line:greedy}), we determine the policy by selecting the action with the highest advantage value based on the learned advantage function.

\begin{algorithm}[!t]
	\caption{\textbf{\actalg}}
	\label{alg:action}
	\begin{algorithmic}[1]
		\State \textbf{Input}: offline datset $\Dcal$.
		\State 	\textbf{MLE}: compute 
  $\textstyle \hA_h = \argmax_{A_h \in \Gcal_{A_h}} \sum_{n=1}^N \log P_{A_h}(o=o_h^{n}\mid s_h^n,a_h^{n,0},a_h^{n,1})$ $\forall h\in[H]$ \label{line:MLE_Q}
		\State \textbf{Greedy policy}: return $\hpi_h(s) = \argmax_{a\in\Acal}\hA_h(s,a)$ \label{line:greedy}

	\end{algorithmic}
\end{algorithm}
\subsection{Analysis}
Now we show that \actalg is able to learn a near-optimal policy as long as offline data covers the optimal policy. Our analysis depends on the following assumption on the margin of $\QS$:
\begin{assumption}[Soft margin]
\label{ass:soft}
There exists $\alpha_0\in\RR^{+}$, $\beta\in(0,\infty]$ such that for all $a\in\Acal,h\in[H],\alpha>0$, we have
$\PP^{\pis,\Pstar}(0<|\QS_h(s_h,\pis(s_h))-\QS_h(s_h,a)|<\alpha)\leq(\alpha/\alpha_0)^{\beta}.$
\end{assumption}

The soft margin is widely used in the literature on classification, decision making, and RL \citep{audibert2007fast,perchet2013multi,luedtke2020performance,hu2021fast,hu2022fast,uehara2023refined}. Note, when the optimal Q function satisfies a gap \citep[as in][]{simchowitz2019non,wu2022gap}, the soft margin assumption holds with $\beta=\infty$.



Next, we introduce the concentrability coefficient for the action-based comparison setting, which is defined as follows.
\begin{definition}[concentrability coefficient for action-based comparison]
\begin{align*}
\Cact:=\sup_{h\in[H],A_h\in\Gcal_{A_h}}\frac{\EE_{(s,a^{0})\sim d^{\pis}_h, a^{1}\sim\Unif(\cdot\mid s)}[l(A_h,s,a^{0},a^{1})]}{\EE_{s\sim\mu_h, a^{0}\sim\mu_{0,h}(\cdot\mid s), a^{1}\sim\mu_{1,h}(\cdot\mid s)}[l(A_h,s,a^{0},a^{1})]},
\end{align*}
where $l(A_h,s,a^{0},a^{1}):=|\AS_h(s,a^{0})-\AS_h(s,a^{1})-A_h(s,a^{0})+A_h(s,a^{1})|^2$ and $\Unif(\cdot\mid s)$ is the uniform policy over $\Acal$.
\end{definition}
We observe that
\begin{align*}
\Cact 
\leq\bigg(\sup_{h\in[H],s\in\Scal}\frac{d^{\pis}_h(s)}{\mu_h(s)}\bigg)\cdot\bigg(\sup_{h\in[H],s\in\Scal,a^0\in\Acal}\frac{\pis_h(a^0\mid s)}{\mu_{0,h}(a^0\mid s)}\bigg)\cdot\bigg(\frac{1}{|\Acal|}\sup_{h\in[H],s\in\Scal,a^1\in\Acal}\frac{1}{\mu_{1,h}(a^1\mid s)}\bigg).
\end{align*} 
Based on this bound, we can consider simple sufficient conditions for $\Cact$ to be finite. Firstly, regarding the first term, it is sufficient for the dataset distribution $\mu_h$ to cover the states visited by the optimal policy $\pis$, denoted as $d^{\pis}_h$. Regarding the second term, we require $\mu_{0,h}$ to cover $\pis_h$. Additionally, the third term can be upper bounded when $\mu_{1,h}$ can cover the whole action space. This is mild because $\forall s\in \Scal;\mu_{h}(s)>0$ is not controllable to the learner; but $\forall (s,a)\in \Scal\times \Acal;\mu_{1,h}(a\mid s)>0$ is controllable to the learner in the data-collection process. To summarize, $\Cact <\infty$ primarily requires partial coverage over the state space with respect to the optimal policy, which is preferable in practical applications where $\Scal$ can be very large.

Additionally, we introduce several assumptions on the function classes similar to those in Section~\ref{sec:known}.
\begin{assumption}
\label{ass:realize3}
For all $h\in[H]$, we have $\AS_h\in\Gcal_{A_h}$.
\end{assumption}

\begin{assumption}
\label{ass:bound2}
For all $h\in[H]$ and $A_h\in\Gcal_{A_h}$, we have $|A_h(s,a)|\leq b_{\max}$ for all $(s,a)\in\Scal\times\Acal$.
\end{assumption}
With the aforementioned assumptions, we can establish the sample complexity of \actalg.
\begin{theorem}
\label{thm:action}
Under Assumption~\ref{ass:preference_action},\ref{ass:soft},\ref{ass:realize3} and \ref{ass:bound2}, we have with probability at least $1-\delta$ that
\begin{align*}
J(\pis;\rstar,\Pstar)-J(\hpi; \rstar, \Pstar)\leq cH|\Acal|\bigg(\frac{2}{\beta}\bigg)^{\frac{\beta-2}{\beta+2}}\bigg(\frac{1}{\alpha_0}\bigg)^{\frac{2\beta}{\beta+2}}\bigg(\frac{\kappa^2_A \Cact\log(H\Ncal_{\Gcal_{A}}(1/N)/\delta)}{N}\bigg)^{\frac{\beta}{\beta+2}},
\end{align*}
where $\Ncal_{\Gcal_{A}}:=\max_{h\in[H]}\Ncal_{\Gcal_{A_h}}$ and $\kappa_A=\frac{1}{\inf_{x\in[-b_{\max},b_{\max}]}\Phi'(x)}$.
\end{theorem}
Theorem~\ref{thm:action} suggests that \actalg can learn a near-optimal policy as long as $\Cact$ takes a finite value under a soft margin. Specifically, when a hard margin is imposed (i.e., $\beta=\infty$), \actalg can learn an $\eps$-optimal policy with a sample complexity of $N=\TO(1/\eps)$, which is faster than a typical rate $\TO(1/\eps^2)$. As mentioned earlier, the quantity $\Cact$ represents the extent to which the distribution induced by the optimal policy is covered by the offline data. Therefore, there is no need for a potentially stringent condition that requires the offline data to cover the entire state space like \citet{zhu2023principled}.

Furthermore, our guarantee is designed to overcome the limitations of existing approaches. In Theorem \ref{thm:main}, our upper-bound is influenced by the parameter $\kappa$. When using a common sigmoid link function, this parameter scales with $\Theta(\exp(r_{\max}))$. As a result, in dense reward settings where $r_{\max}$ scales with $H$, this scaling factor may lead to an explicit dependence of $\Theta(\exp(H))$. Similar observations have been made in previous works \citep{zhu2023principled,pacchiano2021dueling,chen2022human}. However, even if $r_{\max}$ scales with $H$, it is known that the $\ell_{\infty}$-norm of the advantage function, denoted as $b_{\max}$, can take much smaller values \citep{ross2011reduction,agarwal2019reinforcement} Hence, we can avoid the explicit dependence on $\Theta(\exp(H))$.



%% file: summary.tex
\section{Conclusions}
We propose the first algorithm for trajectory-wise PbRL with general function approximation and under partial coverage. We establish lower bounds that explain the differences between our PbRL model and standard RL with direct reward feedback. Moreover, we extend our algorithm to unknown transitions and to preference feedback over actions, all while maintaining strong guarantees under partial coverage.

%% file: discuss.tex
\section{Feasible Implementation of \mainalg}
\label{sec:discussion}
In this section we show how to implement the robust optimization step (Line 4) of \mainalg in practice. Our idea is inspired by standard offline RL \citep{rigter2022rambo} where the authors rely on Lagrangian formulation to make the theoretical algorithm CPPO \citep{uehara2021pessimistic} practical enough to achieve good performance on the D4RL datasets. We believe the empirical insights provided in \citep{rigter2022rambo} can be applied here as well.

First for the Lagrangian relaxation, the original inner minimization problem in Line 4 of \mainalg is 
\begin{align*} 
\min _ {r\in\mathcal{R}(\mathcal{D})} J(\pi;r, P^*) - \mathbb{E} _ {\tau\sim\muref} [r(\tau)]. 
\end{align*} 
Note that the only constraint is $r\in\mathcal{R}(\mathcal{D})$. Then by introducing a Lagrangian multiplier $\beta$, we can convert such constrained minimization problem into an unconstrained regularized minimization problem: \begin{align*} \min _ {r} J(\pi;r, P^*) - \mathbb{E} _ {\tau\sim\muref} [r(\tau)]-\beta \sum_{n=1}^N \log P_{r}(o=o^n|\tau^{n,0}, \tau^{n,1}). \end{align*}

Consequently, Line 4 in \mainalg can be converted to the following unconstrained regularized max-min problem: \begin{align*} \max _ {\pi} \min _ {r} \mathcal{L}(\pi,r):=J(\pi;r, P^*) - \mathbb{E} _ {\tau\sim\muref} [r(\tau)]-\beta \sum_{n=1}^N \log P_{r}(o=o^n|\tau^{n,0}, \tau^{n,1}). \end{align*}

Since now we are facing an unregularized problem, the most common way to solve $\mathcal{L}(\pi,r)$ in practice is gradient ascent-descent. Suppose $\pi$ and $r$ are parametrized by $\theta$ and $\lambda$ (usaully neural networks). Then gradient ascent-descent requires us to compute an unbiased stochastic gradient with respect to $\theta$ and $\lambda$ respectively. Fortunately, this can be easy to achieve in practice. On the one hand, for the gradient of $\theta$, we only need to compute $\nabla_{ \theta} J(\pi _ {\theta}; r, P^*)$. This task has been thoroughly discussed in the literature of policy gradient and one example is REINFORCE, which samples a trajectory $\tau$ by executing $\pi_{\theta}$ in $P^*$ and then the estimated graidient can be expressed as \begin{align*} r(\tau)\sum _ {h=1} ^H \nabla _{\theta} \pi _ {\theta,h} (a_h|s_h), \end{align*} where $(s_h,a_h)$ is the $h$-step of $\tau$.

On the other hand, for the gradient of $\lambda$, we only need to sample independent trajecotories $\tau'$ by executing $\pi_{\theta}$ in $P^*$ and $\tau''$ from $\muref$ and an index $i\in[N]$. Then the unbiased estimated gradient can be directly written as \begin{align*} \nabla_{\lambda} r _ {\lambda}(\tau') - \nabla_{\lambda} r _ {\lambda}(\tau'') - \beta\nabla_{\lambda}\log P_{r_{\lambda}}(o=o^i|\tau^{i,0}, \tau^{i,1}). \end{align*}

Therefore, with the above estimated gradients, we can then run graident ascent-descent happily to solve $\max _ {\pi} \min _ {r} \mathcal{L}(\pi,r)$ in practice.

%% file: appendix_bracket.tex
\section{Proof of Proposition~\ref{prop:bracket-linear}}
\label{proof:prop-bracket-linear}
Let $\Fcal$ denote the function class $\{f_r:f_r(\tau^0,\tau^1)=P_r(o=1|\tau^0,\tau^1),r\in\Gcal_r\}$. Let $\Ical_{\Fcal}(\eps)$ denote the $\eps$-bracket number with respect to $\ell_{\infty}$-norm, i.e., the minimum integer $M$ such that there exist $M$ functions $\{f^i\}_{i=1}^M$ such that for each $f_r\in\Fcal$, we have $\sup_{\tau^0,\tau^1}|f_r(\tau^0,\tau^1)-f^i(\tau^0,\tau^1)|\leq\eps$ for some $i\in[M]$. Then we know there exists a set of function $\overline{\Fcal}$ with $|\overline{\Fcal}|=\Ical_{\Fcal}(\eps/4)$ such that for each $f_r\in\Fcal$, there exists $\of\in\overline{\Fcal}$ satisfying
\begin{align*}
\sup_{\tau^0,\tau^1}|f_r(\tau^0,\tau^1)-\of(\tau^0,\tau^1)|\leq\eps/4.
\end{align*}

Now we construct a bracket $(g^1_{\of},g^2_{\of})$ defined as follows:
\begin{align*}
&g^1_{\of}(o=1|\tau^0,\tau^1)=\of(\tau^0,\tau^1)-\eps/4,g^1_{\of}(o=0|\tau^0,\tau^1)=1-\of(\tau^0,\tau^1)-\eps/4,\\
&g^2_{\of}(o=1|\tau^0,\tau^1)=\of(\tau^0,\tau^1)+\eps/4,g^2_{\of}(o=0|\tau^0,\tau^1)=1-\of(\tau^0,\tau^1)+\eps/4.
\end{align*}
Then clearly we have $g^1_{\of}(\cdot|\tau^0,\tau^1)\leq P_r(\cdot|\tau^0,\tau^1)\leq g^2_{\of}(\cdot|\tau^0,\tau^1)$ and $\Vert g^1_{\of}(\cdot|\tau^0,\tau^1)-g^2_{\of}(\cdot|\tau^0,\tau^1)\Vert_1\leq\eps$. This implies that $\Ncal_{\Gcal_r}(\eps)\leq\Ical_{\Fcal}(\eps/4)$.

Now we only need to bound $\Ical_{\Fcal}(\eps/4)$. Consider $\theta$ and $\theta'$ with
$\Vert\theta-\theta'\Vert_2\leq\eps_{1}$ and let $r$ ($r'$) denote the reward $\langle\phi,\theta\rangle$ ($\langle\phi,\theta'\rangle$). Then we know for all $\tau$,
\begin{align*}
|r(\tau)-r'(\tau)|\leq R\eps_1.
\end{align*}

Fix the trajectory pair $(\tau^0,\tau^1)$. Without loss of generality, we assume $\exp(r(\tau^0))+\exp(r(\tau^1))\leq\exp(r'(\tau^0))+\exp(r'(\tau^1))$. Then we have
\begin{align*}
\exp(r(\tau^0))+\exp(r(\tau^1))\leq\exp(r'(\tau^0))+\exp(r'(\tau^1))\leq\exp(R\eps_1)\Big(\exp(r(\tau^0))+\exp(r(\tau^1))\Big).
\end{align*}

On the other hand, we have 
\begin{align*}
&|f_{r}(\tau^0,\tau^1)-f_{r'}(\tau^0,\tau^1)|\\
&\qquad=\frac{\Big|\exp(r(\tau^1))\Big(\exp(r'(\tau^0))+\exp(r'(\tau^1))\Big)-\exp(r'(\tau^1))\Big(\exp(r(\tau^0))+\exp(r(\tau^1))\Big)\Big|}{\Big(\exp(r'(\tau^0))+\exp(r'(\tau^1))\Big)\Big(\exp(r(\tau^0))+\exp(r(\tau^1))\Big)}.
\end{align*}

Therefore, if $\exp(r(\tau^1))\Big(\exp(r'(\tau^0))+\exp(r'(\tau^1))\Big)-\exp(r'(\tau^1))\Big(\exp(r(\tau^0))+\exp(r(\tau^1))\Big)\geq0$, then we have
\begin{align*}
&\Big|\exp(r(\tau^1))\Big(\exp(r'(\tau^0))+\exp(r'(\tau^1))\Big)-\exp(r'(\tau^1))\Big(\exp(r(\tau^0))+\exp(r(\tau^1))\Big)\Big|\\
\leq&\exp(R\eps_1)\exp(r(\tau^1))\Big(\exp(r(\tau^0))+\exp(r(\tau^1))\Big)-\exp(-R\eps_1)\exp(r(\tau^1))\Big(\exp(r(\tau^0))+\exp(r(\tau^1))\Big)\\
=&(\exp(R\eps_1)-\exp(-R\eps_1))\exp(r(\tau^1))\Big(\exp(r(\tau^0))+\exp(r(\tau^1))\Big).
\end{align*}

Otherwise, we have
\begin{align*}
&\Big|\exp(r(\tau^1))\Big(\exp(r'(\tau^0))+\exp(r'(\tau^1))\Big)-\exp(r'(\tau^1))\Big(\exp(r(\tau^0))+\exp(r(\tau^1))\Big)\Big|\\
\leq&\exp(R\eps_1)\exp(r(\tau^1))\Big(\exp(r(\tau^0))+\exp(r(\tau^1))\Big)-\exp(r(\tau^1))\Big(\exp(r(\tau^0))+\exp(r(\tau^1))\Big)\\
=&(\exp(R\eps_1)-1)\exp(r(\tau^1))\Big(\exp(r(\tau^0))+\exp(r(\tau^1))\Big).
\end{align*}

Therefore we have
\begin{align*}
&|f_{r}(\tau^0,\tau^1)-f_{r'}(\tau^0,\tau^1)|\\
&\qquad\leq\frac{(\exp(R\eps_1)-\exp(-R\eps_1))\exp(r(\tau^1))\Big(\exp(r(\tau^0))+\exp(r(\tau^1))\Big)}{\Big(\exp(r'(\tau^0))+\exp(r'(\tau^1))\Big)\Big(\exp(r(\tau^0))+\exp(r(\tau^1))\Big)}\leq\exp(2R\eps_1)-1.
\end{align*}

This implies that for any $\eps\leq1$,
\begin{align*}
\log\Ical_{\Fcal}(\eps/4)\leq\log\Ical_{d,B}\Big(\frac{2\ln2}{R}\eps\Big)\leq\Ocal\Big(d\log\frac{BR}{\eps}\Big),
\end{align*}
 where $\Ical_{d,B}(\cdot)$ is the covering number of a $d$-dimensional ball centered at the origin with radius $B$ with respect to $\ell_2$-norm and the last step is from \cite{wainwright2019high}. This concludes our proof.

%% file: sketch.tex
\section{Proof of Theorem~\ref{thm:main}} 
The proof of Theorem~\ref{thm:main} consists of two steps, deriving the guarantee of MLE and analyzing the performance of pessimistic offline RL.

\paragraph{Step 1: MLE guarantee.} We first need to show that the confidence set $\Rcal(\Dcal)$ contains the true reward $\rstar$ with high probability. This can be proved via the following lemma which characterizes the guarantee of MLE:
\begin{lemma}[Performance of MLE]
\label{lem:mle}
Fix any $\delta\in(0,1]$. Then with probability at least $1-\delta/2$  we have that for all reward function $r\in\Gcal_r$,
\begin{align*}
\sum_{n=1}^N\log\bigg(\frac{P_r(o^{n}|\tau^{n,0},\tau^{n,1})}{P_{\rstar}(o^{n}|\tau^{n,0},\tau^{n,1})}\bigg)\leq\cmle\log(\Ncal_{\Gcal_r}(1/N)/\delta),
\end{align*}
where $\cmle>0$ is a universal constant.
\end{lemma}
We defer the proof to Appendix~\ref{proof:lem-mle}. Denote the event in Lemma~\ref{lem:mle} by $\Ecal_1$, then we know $\PP(\Ecal_1)\geq 1-\delta/2$. Under the event $\Ecal_1$, we have 
\begin{align*}
\sum_{n=1}^N\log P_{\rstar}(o^{n}|\tau^{n,0},\tau^{n,1})\geq\sum_{n=1}^N\log P_{\hr}(o^{n}|\tau^{n,0},\tau^{n,1})-\cmle\log(\Ncal_{\Gcal_r}(1/N)/\delta),
\end{align*}
which implies that $\rstar\in\Rcal(\Dcal)$ since we know $\rstar\in\Gcal_r$ from Assumption~\ref{ass:realize}.

Nevertheless, the confidence set $\Rcal(\Dcal)$ is constructed via loglikelihood and we indeed prefer a bound on the total variation (TV) distance between $P_{r}$ and $P_{\rstar}$ where $r\in\Rcal(\Dcal)$ to facilitate our subsequent analysis. We can obtain such a bound as shown in the following lemma from the literature (\cite{liu2022partially}[Proposition 14],\cite{zhan2022pac}[Lemma 9]):
\begin{lemma}
\label{lem:tv}
With probability at least $1-\delta/2$, we have for all reward function $r\in\Gcal_r$ that
\begin{align*}
\EE_{\tau^{0}\sim\mu_0,\tau^1\sim\mu_1}\bigg[\Big\Vert P_{r}(\cdot|\tau^0,\tau^1)-P_{\rstar}(\cdot|\tau^0,\tau^1)\Big\Vert_1^2\bigg]\leq \frac{\ctv}{N}\bigg(\sum_{n=1}^N\log\bigg(\frac{P_{\rstar}(o^{n}|\tau^{n,0},\tau^{n,1})}{P_{r}(o^{n}|\tau^{n,0},\tau^{n,1})}\bigg)+\log(\Ncal_{\Gcal_r}(1/N)/\delta)\bigg),
\end{align*}
where $\ctv>0$ is a universal constant.
\end{lemma}

Denote the event in Lemma~\ref{lem:tv} by $\Ecal_2$ and then we know $\PP(\Ecal_2)\geq 1-\delta/2$. Then from Lemma~\ref{lem:mle} and Lemma~\ref{lem:tv} we know that under event $\Ecal_1\cap\Ecal_2$, we have for all $r\in\Rcal(\Dcal)$:
\begin{align}
\label{eq:known-1}
\EE_{\tau^{0}\sim\mu_0,\tau^1\sim\mu_1}\bigg[\Big\Vert P_{r}(\cdot|\tau^0,\tau^1)-P_{\rstar}(\cdot|\tau^0,\tau^1)\Big\Vert_1^2\bigg]\leq\frac{c\log(\Ncal_{\Gcal_r}(1/N)/\delta)}{N},
\end{align}
where $c>0$ is a universal constant.

Then under Assumption~\ref{ass:bound}, we can apply the mean value theorem between $r^{\star}(\tau_1)-r^{\star}(\tau_0)$ and $r(\tau_1)-r(\tau_0)$ to \eqref{eq:known-1} and ensure for all $r\in\Rcal(\Dcal)$ that
\begin{align}
\label{eq:known-2}
    \EE_{\tau^0\sim\mu_0,\tau^1\sim\mu_1}[|(r^{\star}(\tau_1)-r^{\star}(\tau_0)) - (r(\tau_1)-r(\tau_0))|^2]\leq \frac{c\kappa^2\log(\Ncal_{\Gcal_r}(1/N)/\delta)}{N},
\end{align}
where $\kappa:=\frac{1}{\inf_{x\in[-\rmax,\rmax]}\Phi'(x)}$ measures the non-linearity of the link function $\Phi$.

\paragraph{Step 2: Pessimistic offline RL.} Let $\rinf_{\pi}$ denote $\argmin_{r\in\Rcal(\Dcal)} J(\pi;r,\Pstar)-\EE_{\tau\sim\muref}[r(\tau)]$. Then we can bound the suboptimality of $\hpi$ as follows:
\begin{align*}
& J(\pit;\rstar,\Pstar)-J(\hpi; \rstar, \Pstar)\\
= & \big(J(\pit;\rstar,\Pstar)-\EE_{\tau\sim\muref}[\rstar(\tau)]\big)-\big(J(\hpi; \rstar, \Pstar)-\EE_{\tau\sim\muref}[\rstar(\tau)]\big)\\
\leq & \Big(\big(J(\pit;\rstar,\Pstar)-\EE_{\tau\sim\muref}[\rstar(\tau)]\big)-\big(J(\pit;\rinf_{\pit},\Pstar)-\EE_{\tau\sim\muref}[\rinf_{\pit}(\tau)]\big)\Big)\\
& - \Big(\big(J(\hpi; \rstar, \Pstar)-\EE_{\tau\sim\muref}[\rstar(\tau)]\big)-\big(J(\hpi;\rinf_{\hpi},\Pstar)-\EE_{\tau\sim\muref}[\rinf_{\hpi}(\tau)]\big)\Big)\\
\leq &\big(J(\pit;\rstar,\Pstar)-\EE_{\tau\sim\muref}[\rstar(\tau)]\big)-\big(J(\pit;\rinf_{\pit},\Pstar)-\EE_{\tau\sim\muref}[\rinf_{\pit}(\tau)]\big)\\    
= & \EE_{\tau^{0}\sim\pit,\tau^1\sim\muref}[(\rstar(\tau^0)-\rstar(\tau^1))-(\rinf_{\pit}(\tau^0)-\rinf_{\pit}(\tau^1))]\\
\leq & C_{r}(\Gcal_r,\pit,\muref)\sqrt{\EE_{\tau_0\sim\mu_0,\tau_1\sim\mu_1}[|\rstar(\tau^0)-\rstar(\tau^1) - \rinf_{\pit}(\tau^0)+\rinf_{\pit}(\tau^1)|^2]}\\
\leq & \sqrt{\frac{cC^2_{r}(\Gcal_r,\pit,\muref)\kappa^2\log(\Ncal_{\Gcal_r}(1/N)/\delta)}{N}},
\end{align*}
where the second step is due to $\hpi=\argmax_{\pi\in\Pihis }\min_{r \in \Rcal(\Dcal)} J(\pi; r,\Pstar)-\EE_{\tau\sim\muref}[r(\tau)]$, the third step is due to $\rinf_{\hpi}=\argmin_{r\in\Rcal(\Dcal)} J(\hpi;r,\Pstar)-\EE_{\tau\sim\muref}[r(\tau)]$, the fifth step comes from the definition of $C_r(\Gcal_r,\pit,\muref)$ (Definition~\ref{def:concentrability}) and the last step leverages \eqref{eq:known-2}. This concludes our proof.

\subsection{Proof of Lemma~\ref{lem:mle}}
\label{proof:lem-mle}
The proof largely follows \cite{zhan2022pac}. Suppose $\BF$ is a $1/N$-bracket of $\Gcal_r$ with $|\BF|=\Ncal_{\Gcal_r}(1/N)$ and we denote the set of all right brackets in $\BF$ by $\TF$, i.e., $\TF:=\{f:\exists f',\text{ such that }[f',f]\in\BF\}$. Then fix any $f\in\TF$, we have:
\begin{align*}
	&\E\bigg[\exp\bigg(\sum_{n=1}^N\log\bigg(\frac{f(o^n|\tau^{n,0},\tau^{n,1})}{P_{\rstar}(o^n|\tau^{n,0},\tau^{n,1})}\bigg)\bigg)\bigg]=\prod_{n=1}^N\E\bigg[\exp\bigg(\log\bigg(\frac{f(o^n|\tau^{n,0},\tau^{n,1})}{P_{\rstar}(o^n|\tau^{n,0},\tau^{n,1})}\bigg)\bigg)\bigg]\\
	&\qquad=\prod_{n=1}^N\E\bigg[\frac{f(o^n|\tau^{n,0},\tau^{n,1})}{P_{\rstar}(o^n|\tau^{n,0},\tau^{n,1})}\bigg]=\prod_{n=1}^N\E\bigg[\sum_{o}f(o|\tau^{n,0},\tau^{n,1})\bigg]\leq\bigg(1+\frac{1}{N}\bigg)^{N}\leq e,
\end{align*}
where the first step is due to each sample in $\Dcal$ is i.i.d., the third step uses Tower property and the fourth step is from the fact that $\BF$ is a minimum $1/N$-bracket.

Then by Markov's inequality we have for any $\delta\in(0,1]$,
\begin{align*}
	&\PP\bigg(\sum_{n=1}^N\log\bigg(\frac{f(o^n|\tau^{n,0},\tau^{n,1})}{P_{\rstar}(o^n|\tau^{n,0},\tau^{n,1})}\bigg)>\log(1/\delta)\bigg)\\
	&\qquad\leq\E\bigg[\exp\bigg(\sum_{n=1}^N\log\bigg(\frac{f(o^n|\tau^{n,0},\tau^{n,1})}{P_{\rstar}(o^n|\tau^{n,0},\tau^{n,1})}\bigg)\bigg)\bigg]\cdot\exp[-\log(1/\delta)]\leq e\delta.
\end{align*} 

By union bound, we have for all $f\in\TF$,
\begin{align*}
	&\PP\bigg(\sum_{n=1}^N\log\bigg(\frac{f(o^n|\tau^{n,0},\tau^{n,1})}{P_{\rstar}(o^n|\tau^{n,0},\tau^{n,1})}\bigg)>\cmle\log(\Ncal_{\Gcal_r}(1/N)/\delta)\bigg)\leq \delta/2,
\end{align*} 
where $\cmle>0$ is a universal constant.

Therefore from the definition of $1/N$-bracket net, we know for all $r\in\Gcal_r$, there exists $f\in\TF$ such that $P_{r}(\cdot|\tau^0,\tau^1)\leq f(\cdot|\tau^0,\tau^1)$ for any trajectories $(\tau^0,\tau^1)$. This implies that for all $r\in\Gcal_r$,
\begin{align*}
	&\PP\bigg(\sum_{n=1}^N\log\bigg(\frac{P_r(o^n|\tau^{n,0},\tau^{n,1})}{P_{\rstar}(o^n|\tau^{n,0},\tau^{n,1})}\bigg)>\cmle\log(\Ncal_{\Gcal_r}(1/N)/\delta)\bigg)\leq \delta/2,
\end{align*}
This concludes our proof.

%% file: proof_lower.tex
\section{Proofs of Lower Bounds}



\subsection{Proof of Proposition~\ref{prop:tr}}
\label{proof:prop-tr}
Given any $S, A,H$,  consider a MDP with horizon $H$, state space $\Scal=\{s^1,s^2,\cdots,s^{S}\}$ and action space $\Acal=\{a^1,a^2,\cdots,a^A\}$. In the following discussion we consider the case $C\geq 2$ and $1<C<2$ respectively. 

\paragraph{Case 1: $C\geq 2$.} Consider the case where the state is fixed throughout an episode.  We suppose the initial state distribution $\rhos$ is $\rhos(s^1)=\frac{1}{2}$ and $\rhos(s^i)=\frac{1}{2(S-1)}$ for all $2\leq i \leq S$. Let $\pi_{\mathrm{tar},h}(a^1|s)=1$ for all $h\in[H]$ and $s\in\Scal$. Then we can set the dataset distribution $\mu_0$ as
\begin{align*}
\mu_0(\tau)=
\begin{cases}
\frac{1}{2C}, &\text{ if the state is $s^1$ and all actions in $\tau$ are $a^1$ except $a_{H-1}=a^2$,}\\
\frac{1}{2}-\frac{1}{2C}, &\text{ if the state is $s^1$ and all actions in $\tau$ are $a^1$ except $a_{H}=a^2$,}\\
\frac{1}{2(S-1)}, &\text{ if the state is not $s^1$ and all actions in $\tau$ are $a^1$,}\\
0, &\text{ otherwise,}
\end{cases}
\end{align*}
where $a_h$ is the action at step $h$ in $\tau$. Then we know 
\begin{align*}
&\mu_{0,h}(s,a^1)=\frac{1}{2(S-1)},\qquad\forall h\in[H], s\in\Scal\setminus\{s^1\},\\
&\mu_{0,h}(s^1,a^1)=\frac{1}{2},\quad\mu_{0,H-1}(s,a^1)= \frac{1}{2}-\frac{1}{2C},\quad\mu_{0,H}(s,a^1)= \frac{1}{2C}.
\end{align*}
It is obvious we have $\Cst\leq C$ in this setting. On the other hand, since the trajectory whose state is $s^1$ and all actions are $a^1$ is covered by $\pit$ but not by $\mu_0$, we have $\Ctr=\infty$.
  
\paragraph{Case 2: $1<C<2$.} Consider the case where the state is fixed throughout an episode. We suppose the initial state distribution of $\rhos$ is $\rhos(s^1)=\frac{C-1}{2}$, $\rhos(s^2)=\frac{2-C}{2}$ and $\rhos(s^i)=\frac{1}{2(S-2)}$ for all $3\leq i \leq S$. Note that here we require $S\geq 3$. When $S=2$, we can let $\rhos(s^1)=C-1$ and $\rhos(s^2)=2-C$ and the following analysis will still hold. Therefore here we assume $S\geq 3$ without loss of generality. Let $\pi_{\mathrm{tar},h}(a^1|s)=1$ for all $h\in[H]$ and $s\in\Scal$. Then we can set the dataset distribution $\mu_0$ as
\begin{align*}
\mu_{0}(\tau)=
\begin{cases}
\frac{C-1}{2C},&\text{ if the state of $\tau$ is $s^1$ and all actions in $\tau$ are $a^1$ except $a_{H-1}=a^2$,}\\
\frac{C-1}{2C},&\text{ if the state of $\tau$ is $s^1$ and all actions in $\tau$ are $a^1$ except $a_{H}=a^2$,}\\
\frac{2-C}{2C},&\text{ if the state of $\tau$ is $s^2$ and the actions are all $a^1$,}\\
\frac{1}{2(S-2)},&\text{ if the state of $\tau$ is not $s^1$ or $s^2$ and the actions are all $a^1$,}\\
0, &\text{ otherwise.}
\end{cases}
\end{align*}
Then we know
\begin{align*}
&\mu_{0,h}(s,a^1)=\frac{1}{2(S-2)}, \qquad\forall h\in[H], s\in\Scal\setminus\{s^1,s^2\},\\
&\mu_{0,h}(s^2,a^1)=\frac{2-C}{2C}, \qquad\forall h\in[H],\\
&\mu_{0,h}(s^1,a^1)=\frac{2C-2}{2C},\qquad\forall h\in[H-2],\\
&\mu_{0,H-1}(s^1,a^1)=\mu_{0,H}(s^1,a^1)=\frac{C-1}{2C}.\\
\end{align*}
It is obvious we have $\Cst\leq C$ in this setting. On the other hand, since the trajectory whose state is $s^1$ and all actions are $a^1$ is covered by $\pit$ but not by $\mu_0$, we have $\Ctr=\infty$. This concludes our proof.

\subsection{Proof of Theorem~\ref{thm:lower}}
\label{proof:thm-lower}
We consider the case $C\geq 2$ and $1<C<2$ respectively.
\paragraph{Case 1: $C\geq 2$.} Consider the case where there is only one state $s$ and two actions $a^1,a^2$. Set the dataset distribution $\mu_0=\mu_1$ where
\begin{align*}
\mu_0(\tau)=
\begin{cases}
\frac{1}{C}, &\text{ if all actions in $\tau$ are $a^1$ except $a_{H-1}=a^2$,}\\
1-\frac{1}{C}, &\text{ if all actions in $\tau$ are $a^1$ except $a_{H}=a^2$,}\\
0, &\text{ otherwise,}
\end{cases}
\end{align*}
where $a_h$ is the action at step $h$ in $\tau$. In the following discussion we will use $\tau^1$ to denote the trajectory where all actions are $a^1$ except $a_{H-1}=a^2$ and $\tau^2$ to denote the trajectory where all actions are $a^1$ except $a_{H-1}=a^2$. Then we know 
\begin{align*}
&\mu_{0,h}(s,a^1)=1,\qquad\forall h\in[H-2],\\
&\mu_{0,H-1}(s,a^1)= 1-\frac{1}{C},\qquad \mu_{0,H-1}(s,a^2)= \frac{1}{C},\\
&\mu_{0,H}(s,a^1)= \frac{1}{C},\qquad\mu_{0,H}(s,a^2)= 1-\frac{1}{C}.
\end{align*}

We consider two different reward function $r^1$ and $r^2$:
\begin{align*}
&r^1_h(s,a^1)=r^1_h(s,a^2)=r^2_h(s,a^1)=r^2_h(s,a^2)=0,\qquad\forall h\in[H-2],\\
&r^1_{H-1}(s,a^1)=r^2_{H-1}(s,a^2)=1,\qquad r^1_{H-1}(s,a^2)=r^2_{H-1}(s,a^1)=0,\\
&r^1_{H}(s,a^1)=r^2_{H}(s,a^2)=1,\qquad r^1_{H}(s,a^2)=r^2_{H}(s,a^1)=0,\\
\end{align*}
Then we have two MDPs, $\Mcal_1$ and $\Mcal_2$ whose reward functions are $r^1$ and $r^2$ respectively. It can be easily verified that $(\Mcal_1,\mu_0)\in\overline{\Theta}_{\mathrm{st}}(C),(\Mcal_2,\mu_0)\in\overline{\Theta}_{\mathrm{st}}(C)$.  

Further, let $L(\pi;\Mcal)$ denote the suboptimality of policy $\pi$ in $\Mcal$, then we have for all policies $\pi$,
\begin{align*}
L(\pi;\Mcal_1)+L(\pi;\Mcal_2)\geq 2.
\end{align*}
Now we can apply Le Cam’s method, which leads to the following inequality
\begin{align*}
\inf_{\hpi}\sup_{\Mcal\in\{\Mcal_1,\Mcal_2\}} \EE_{\Dcal}[L(\pi,\Mcal)]\geq\frac{1}{2}\exp(-N\KL\Big(\mu_0\otimes\mu_1\otimes P_{r^1}\Vert\mu_0\otimes\mu_1\otimes P_{r^2}\Big)).
\end{align*}
It can be observed that $\KL\Big(\mu_0\otimes\mu_1\otimes P_{r^1}\Vert\mu_0\otimes\mu_1\otimes P_{r^2}\Big)=0$ since $r^1(\tau^1)=r^1(\tau^2)=r^2(\tau^1)=r^2(\tau^2)=1$. Therefore we have
\begin{align*}
\inf_{\hpi}\sup_{\Mcal\in\{\Mcal_1,\Mcal_2\}} \EE_{\Dcal}[L(\pi,\Mcal)]\geq\frac{1}{2}.
\end{align*}

\paragraph{Case 2: $1<C<2$.} Consider the case where there are two one states $s^1,s^2$ and two actions $a^1,a^2$. We suppose the initial state distribution of $\rhos$ is fixed as $\rhos(s^1)=C-1$ and $\rhos(s^2)=2-C$. In addition, the state will stay the same throughout the whole episode. Then we can set the dataset distribution $\mu_0=\mu_1$ where
\begin{align*}
\mu_{0}(\tau)=
\begin{cases}
\frac{C-1}{C},&\text{ if the state of $\tau$ is $s^1$ and all actions in $\tau$ are $a^1$ except $a_{H-1}=a^2$,}\\
\frac{C-1}{C},&\text{ if the state of $\tau$ is $s^1$ and all actions in $\tau$ are $a^1$ except $a_{H}=a^2$,}\\
\frac{2-C}{C},&\text{ if the state of $\tau$ is $s^2$ and the actions are all $a^1$,}\\
0, &\text{ otherwise.}
\end{cases}
\end{align*}
In the following discussion we will use $\tau^3$ to denote the trajectory where state is $s^1$ and all actions are $a^1$ except $a_{H-1}=a^2$; $\tau^4$ to denote the trajectory where state is $s^1$ and all actions are $a^1$ except $a_{H-1}=a^2$; $\tau^5$ to denote the trajectory where state is $s^2$ and all actions are $a^1$. Then we know
\begin{align*}
&\mu_{0,h}(s^2,a^1)=\frac{2-C}{C},\qquad\forall h\in[H],\\
&\mu_{0,h}(s^1,a^1)=\frac{2C-2}{C},\qquad\forall h\in[H-2],\\
&\mu_{0,H-1}(s^1,a^1)=\mu_{0,H-1}(s^1,a^2)=\frac{C-1}{C},\\
&\mu_{0,H}(s^1,a^1)=\mu_{0,H}(s^1,a^2)=\frac{C-1}{C}.
\end{align*}

We consider two different reward function $r^1$ and $r^2$:
\begin{align*}
&r^1_h(s^1,a^1)=r^1_h(s^1,a^2)=r^2_h(s^1,a^1)=r^2_h(s^1,a^2)=0,\qquad\forall h\in[H-2],\\
&r^1_{H-1}(s^1,a^1)=r^2_{H-1}(s^1,a^2)=1,\qquad r^1_{H-1}(s^1,a^2)=r^2_{H-1}(s^1,a^1)=0,\\
&r^1_{H}(s^1,a^1)=r^2_{H}(s^1,a^2)=1,\qquad r^1_{H}(s^1,a^2)=r^2_{H}(s^1,a^1)=0,\\
&r^1_{h}(s^2,a^1)=r^1_{h}(s^2,a^2)=r^2_{h}(s^2,a^1)=r^2_{h}(s^2,a^2)=0,\qquad\forall h\in[H-1]\\
&r^1_{H}(s^2,a^1)=r^2_{H}(s^2,a^1)=1,\qquad r^1_{H}(s^2,a^2)=r^2_{H}(s^2,a^2)=0.\\
\end{align*}
Then we have two MDPs, $\Mcal_1$ and $\Mcal_2$ whose reward functions are $r^1$ and $r^2$ respectively. It can be easily verified that $(\Mcal_1,\mu_0)\in\overline{\Theta}_{\mathrm{st}}(C),(\Mcal_2,\mu_0)\in\Theta_{\mathrm{st}}(C)$.  

In addition, we have for all policies $\pi$,
\begin{align*}
L(\pi;\Mcal_1)+L(\pi;\Mcal_2)\geq 2(C-1).
\end{align*}
Therefore by Le Cam's method, we have
\begin{align*}
\inf_{\hpi}\sup_{\Mcal\in\{\Mcal_1,\Mcal_2\}} \EE_{\Dcal}[L(\pi,\Mcal)]\geq\frac{(C-1)}{2}\exp\bigg(-N\cdot\KL\Big(\mu_0\otimes\mu_1\otimes P_{r^1}\Vert\mu_0\otimes\mu_1\otimes P_{r^2}\Big)\bigg),
\end{align*}
where the KL divergence is 0 since $r(\tau)=1$ for all $r\in\{r^1,r^2\}$ and $\tau\in\{\tau^3,\tau^4,\tau^5\}$. Therefore, we have
\begin{align*}
\inf_{\hpi}\sup_{\Mcal\in\{\Mcal_1,\Mcal_2\}} \EE_{\Dcal}[L(\pi,\Mcal)]\geq \frac{C-1}{2}.
\end{align*}

In conclusion, we have for any $C>1$ and $H\geq 2$,
\begin{align*}
\inf_{\hpi}\sup_{(\Mcal,\mu_0)\in\Theta_{\mathrm{st}}(C)}\EE_{\Dcal}[J(\pis;\rstar,\Pstar)-J(\hpi;\rstar,\Pstar)]\gtrsim\min\bigg\{C-1,1\bigg\}.
\end{align*}

\subsection{Proof of Theorem~\ref{thm:lower2}}
\label{proof:thm-lower2}
The proof is inspired by the hard instances in \cite{rashidinejad2021bridging}. We consider the case $C\geq 2$ and $1<C<2$ respectively.

\paragraph{Case 1: $C\geq 2$.} Consider the case where there is only one state $s$ and two actions $a^1,a^2$. Set the dataset distribution $\mu_0=\mu_1$ where
\begin{align*}
\mu_{0}(\tau^{\star})=\frac{1}{C},\qquad\mu_{0}(\tau^\dagger)=1-\frac{1}{C},
\end{align*}
where $\tau^{\star}$ is the trajecotry where the actions are all $a^1$ and $\tau^{\dagger}$ is the trajecotry where the actions are all $a^2$.

We consider two different reward function $r^1$ and $r^2$:
\begin{align*}
r^1(\tau)=
\begin{cases}
\frac{1}{2}+x, &\text{ if all the actions in $\tau$ are $a^1$,}\\
\frac{1}{2}, &\text{ otherwise.}
\end{cases}\\
r^2(\tau)=
\begin{cases}
\frac{1}{2}-x, &\text{ if all the actions in $\tau$ are $a^1$,}\\
\frac{1}{2}, &\text{ otherwise.}
\end{cases}
\end{align*}
Here $0<x<\frac{1}{2}$ is a quantity we will specify later. Then we have two MDPs, $\Mcal_1$ and $\Mcal_2$ whose reward functions are $r^1$ and $r^2$ respectively. It can be easily verified that $(\Mcal_1,\mu_0)\in\Theta_{\mathrm{tr}}(C),(\Mcal_2,\mu_0)\in\Theta_{\mathrm{tr}}(C)$.  

Further, let $L(\pi;\Mcal)$ denote the suboptimality of policy $\pi$ in $\Mcal$, then we have for all policies $\pi$,
\begin{align*}
L(\pi;\Mcal_1)+L(\pi;\Mcal_2)\geq x.
\end{align*}
Now we can apply Le Cam's method, which leads to the following inequality
\begin{align*}
\inf_{\hpi}\sup_{\Mcal\in\{\Mcal_1,\Mcal_2\}} \EE_{\Dcal}[L(\pi,\Mcal)]\geq\frac{x}{4}\exp\bigg(-N\cdot\KL\Big(\mu_0\otimes\mu_1\otimes P_{r^1}\Vert\mu_0\otimes\mu_1\otimes P_{r^2}\Big)\bigg).
\end{align*}
Now we only need to bound $\KL\Big(\mu_0\otimes\mu_1\otimes P_{r^1}\Vert\mu_0\otimes\mu_1\otimes P_{r^2}\Big)$, which can be computed as follows:
\begin{align*}
&\KL\Big(\mu_0\otimes\mu_1\otimes P_{r^1}\Vert\mu_0\otimes\mu_1\otimes P_{r^2}\Big)\\
=&2\sum_{\tau^0=\tau^\star,\tau^1=\tau^\dagger}\mu_0(\tau^0)\mu_1(\tau^1)\KL\big(\Bern(\sigma(x))\Vert\Bern(\sigma(-x))\big)\\
\leq& \frac{2\exp(1/2)x^2}{C}.
\end{align*}
Then by letting $x=\min\bigg\{\frac{1}{2},\sqrt{\frac{C}{2\exp(1/2)N}}\bigg\}$, we have
\begin{align*}
\inf_{\hpi}\sup_{\Mcal\in\{\Mcal_1,\Mcal_2\}} \EE_{\Dcal}[L(\pi,\Mcal)]\geq \frac{\exp(-1)}{4}x=\frac{\exp(-1)}{4}\min\bigg\{\frac{1}{2},\sqrt{\frac{C}{2\exp(1/2)N}}\bigg\}.
\end{align*}

\paragraph{Case 2: $1<C<2$.} Consider the case where there are two one states $s^1,s^2$ and two actions $a^1,a^2$. We suppose the initial state distribution of $\rhos$ is fixed as $\rhos(s^1)=C-1$ and $\rhos(s^2)=2-C$. In addition, the state will stay the same throughout the whole episode. Then we can set the dataset distribution $\mu_0=\mu_1$ where
\begin{align*}
\mu_{0}(\tau)=
\begin{cases}
\frac{2(C-1)}{C}\cdot\frac{1}{2},&\text{ if the state of $\tau$ is $s^1$ and the actions are all $a^1$ or all $a^2$,}\\
\frac{2-C}{C},&\text{ if the state of $\tau$ is $s^2$ and the actions are all $a^1$,}\\
0, &\text{ if the state of $\tau$ is $s^2$ and the actions contain $a^2$.}
\end{cases}
\end{align*}
Let $\tau^{\star}$ be the trajectory where the state is $s^1$ and the actions are all $a^1$.

We further consider two different reward function $r^1$ and $r^2$:
\begin{align*}
r^1(\tau)=
\begin{cases}
\frac{1}{2}+x, &\text{ if the state is $s^1$ and all the actions in $\tau$ are $a^1$,}\\
\frac{1}{2}, &\text{ otherwise.}
\end{cases}\\
r^2(\tau)=
\begin{cases}
\frac{1}{2}-x, &\text{ if the state is $s^1$ and all the actions in $\tau$ are $a^1$,}\\
\frac{1}{2}, &\text{ otherwise.}
\end{cases}
\end{align*}
Here $0<x<\frac{1}{2}$ is a quantity we will specify later. Then we have two MDPs, $\Mcal_1$ and $\Mcal_2$ whose reward functions are $r^1$ and $r^2$ respectively. It can be easily verified that $(\Mcal_1,\mu_0)\in\Theta_{\mathrm{tr}}(C),(\Mcal_2,\mu_0)\in\Theta_{\mathrm{tr}}(C)$.  

In addition, we have for all policies $\pi$,
\begin{align*}
L(\pi;\Mcal_1)+L(\pi;\Mcal_2)\geq (C-1)x.
\end{align*}
Therefore by Le Cam's method, we have
\begin{align*}
\inf_{\hpi}\sup_{\Mcal\in\{\Mcal_1,\Mcal_2\}} \EE_{\Dcal}[L(\pi,\Mcal)]\geq\frac{(C-1)x}{4}\exp\bigg(-N\cdot\KL\Big(\mu_0\otimes\mu_1\otimes P_{r^1}\Vert\mu_0\otimes\mu_1\otimes P_{r^2}\Big)\bigg),
\end{align*}
where the KL divergence can be computed as follows:
\begin{align*}
&\KL\Big(\mu_0\otimes\mu_1\otimes P_{r^1}\Vert\mu_0\otimes\mu_1\otimes P_{r^2}\Big)\\
=&2\sum_{\tau^0=\tau^\star,\tau^1\neq\tau^\star}\mu_0(\tau^0)\mu_1(\tau^1)\KL\big(\Bern(\sigma(x))\Vert\Bern(\sigma(-x))\big)\\
\leq& \frac{2(C-1)\exp(1/2)x^2}{C}.
\end{align*}
Then by letting $x=\min\bigg\{\frac{1}{2},\sqrt{\frac{C}{2\exp(1/2)(C-1)N}}\bigg\}$, we have
\begin{align*}
\inf_{\hpi}\sup_{\Mcal\in\{\Mcal_1,\Mcal_2\}} \EE_{\Dcal}[L(\pi,\Mcal)]\geq \frac{(C-1)\exp(-1)}{4}x=\frac{\exp(-1)}{4}\min\bigg\{\frac{C-1}{2},\sqrt{\frac{(C-1)}{2\exp(1/2)N}}\bigg\}.
\end{align*}


In conclusion, we have for any $C>1$ and $H\geq 1$,
\begin{align*}
\inf_{\hpi}\sup_{(\Mcal,\mu_0)\in\Theta_{\mathrm{st}}(C)}\EE_{\Dcal}[J(\pis;\rstar,\Pstar)-J(\hpi;\rstar,\Pstar)]\gtrsim\min\bigg\{C-1,\sqrt{\frac{C-1}{N}}\bigg\}.
\end{align*}

%% file: proof_unknown.tex
\section{Proof of Theorem~\ref{thm:unknown}}
\label{proof:thm-unknown}
The proof still consists of two steps, deriving the guarantee of MLE and analyzing the performance of pessimistic offline RL.

\paragraph{Step 1: MLE guarantee.} Note that Lemma~\ref{lem:mle} and Lemma~\ref{lem:tv} still applies here. Let $\Ecal_1$ and $\Ecal_2$ denote the event in Lemma~\ref{lem:mle} and Lemma~\ref{lem:tv} respectively. Following almost the same arguments, we have the following guarantee for the estimation of the system dynamics:
\begin{lemma}
\label{lem:system}
Under Assumption~\ref{ass:realize2}, with probability at least $1-\delta/2$, the following event holds true:
\begin{align*}
&(1)\Pstar_h\in\Pcal_h(\Dcal),\rhos\in\Pini(\Dcal),\qquad\forall h\in[H-1],\\
&(2)\EE_{(s_h,a_h)\sim\mu_{0,h}}\bigg[\Big\Vert P_h(\cdot|s,a)-\Pstar_h(\cdot|s,a)\Big\Vert_1^2\bigg]+\EE_{(s_h,a_h)\sim\mu_{1,h}}\bigg[\Big\Vert P_h(\cdot|s,a)-\Pstar_h(\cdot|s,a)\Big\Vert_1^2\bigg]\\
&\qquad\leq\frac{c\log(H\Ncal_{\Gcal_{P_h}}(1/N)/\delta)}{N},\qquad\forall h\in[H-1], P_h\in\Pcal_h(\Dcal),\\
&(3)\EE_{s\sim\mu_{0,1}}\bigg[\Big\Vert \rho(s)-\rhos(s)\Big\Vert_1^2\bigg]+\EE_{s\sim\mu_{1,1}}\bigg[\Big\Vert \rho(s)-\rhos(s)\Big\Vert_1^2\bigg]\\
&\qquad\leq\frac{c\log(H\Ncal_{\Gcal_{\rho}}(1/N)/\delta)}{N},\qquad\forall \rho\in\Pcal_0(\Dcal).
\end{align*}
\end{lemma}
The proof is omitted here. Let $\Ecal_3$ denote the event in Lemma~\ref{lem:system}.

\paragraph{Step 2: Pessimistic offline RL.} We first introduce the following lemma which suggests that under event $\Ecal_3$, we can evaluate the expected cumulative reward of $\pit$ with respect to any reward function $r\in\Gcal_r$ via the system dynamics $P_h\in\Pcal_h(\Dcal)$:
\begin{lemma}
\label{lem:performance}
Suppose Asusmption~\ref{ass:bound} is true. Then under $\Ecal_3$, we have for all reward function $r\in\Gcal_r$ and $P=(\{P_h\}_{h=0}^{H-1})$ where $P_h\in\Pcal_h(\Dcal)$ that
\begin{align*}
J(\pit;r,\Pstar)-J(\pit;r,P)\leq H\rmax\sqrt{\frac{cC^2_P(\{\Gcal_{P_h}\},\pit)\log(H\Ncal_{P}(1/N)/\delta)}{N}},
\end{align*}
where $\Ncal_P=\max_{0\leq h\leq H-1}\{\Ncal_{\Gcal_{P_h}}\}$.
\end{lemma}
The proof is deferred to Appendix~\ref{proof:lem-performance}. 

Let $(\rinf_{\pi},\Pinf_{\pi})$ denote $\argmin_{r\in\Rcal(\Dcal),P\in\Pini(\Dcal)\times\prod_{h=1}^{H-1}\Pcal_{h}(\Dcal)} J(\pi;r,P)-\EE_{\tau\sim\muref}[r(\tau)]$. Then under the event $\Ecal_{3}$, we can bound the suboptimality of $\hpi$ as follows:
\begin{align*}
& J(\pit;\rstar,\Pstar)-J(\hpi; \rstar, \Pstar)\\
= & \big(J(\pit;\rstar,\Pstar)-\EE_{\tau\sim\muref}[\rstar(\tau)]\big)-\big(J(\hpi; \rstar, \Pstar)-\EE_{\tau\sim\muref}[\rstar(\tau)]\big)\\
= & \Big(\big(J(\pit;\rstar,\Pstar)-\EE_{\tau\sim\muref}[\rstar(\tau)]\big)-\big(J(\pit;\rinf_{\pit},\Pstar)-\EE_{\tau\sim\muref}[\rinf_{\pit}(\tau)]\big)\Big)\\
& + \Big(\big(J(\pit;\rinf_{\pit},\Pstar)-\EE_{\tau\sim\muref}[\rinf_{\pit}(\tau)]\big)-\big(J(\pit;\rinf_{\pit},\Pinf_{\pit})-\EE_{\tau\sim\muref}[\rinf_{\pit}(\tau)]\big)\Big)\\
& + \Big(\big(J(\pit;\rinf_{\pit},\Pinf_{\pit})-\EE_{\tau\sim\muref}[\rinf_{\pit}(\tau)]\big)-\big(J(\hpi;\rinf_{\hpi},\Pinf_{\hpi})-\EE_{\tau\sim\muref}[\rinf_{\hpi}(\tau)]\big)\Big)\\
& + \Big(\big(J(\hpi;\rinf_{\hpi},\Pinf_{\hpi})-\EE_{\tau\sim\muref}[\rinf_{\hpi}(\tau)]\big)-\big(J(\hpi; \rstar, \Pstar)-\EE_{\tau\sim\muref}[\rstar(\tau)]\big)\Big)\\
\leq & \Big(\big(J(\pit;\rstar,\Pstar)-\EE_{\tau\sim\muref}[\rstar(\tau)]\big)-\big(J(\pit;\rinf_{\pit},\Pstar)-\EE_{\tau\sim\muref}[\rinf_{\pit}(\tau)]\big)\Big)\\
& + \Big(\big(J(\pit;\rinf_{\pit},\Pstar)-\EE_{\tau\sim\muref}[\rinf_{\pit}(\tau)]\big)-\big(J(\pit;\rinf_{\pit},\Pinf_{\pit})-\EE_{\tau\sim\muref}[\rinf_{\pit}(\tau)]\big)\Big)\\
& + \Big(\big(J(\hpi;\rinf_{\hpi},\Pinf_{\hpi})-\EE_{\tau\sim\muref}[\rinf_{\hpi}(\tau)]\big)-\big(J(\hpi; \rstar, \Pstar)-\EE_{\tau\sim\muref}[\rstar(\tau)]\big)\Big)\\
\leq & \Big(\big(J(\pit;\rstar,\Pstar)-\EE_{\tau\sim\muref}[\rstar(\tau)]\big)-\big(J(\pit;\rinf_{\pit},\Pstar)-\EE_{\tau\sim\muref}[\rinf_{\pit}(\tau)]\big)\Big)\\
& + \Big(\big(J(\pit;\rinf_{\pit},\Pstar)-\EE_{\tau\sim\muref}[\rinf_{\pit}(\tau)]\big)-\big(J(\pit;\rinf_{\pit},\Pinf_{\pit})-\EE_{\tau\sim\muref}[\rinf_{\pit}(\tau)]\big)\Big)\\
\leq & \sqrt{\frac{cC^2_{r}(\Gcal_r,\pit,\muref)\kappa^2\log(\Ncal_{\Gcal_r}(1/N)/\delta)}{N}}+H\rmax\sqrt{\frac{cC^2_P(\{\Gcal_{P_h}\},\pit)\log(H\Ncal_{P}(1/N)/\delta)}{N}},
\end{align*}
where the third and fourth step are due to the definition of $\hpi,(\rinf_{\hpi},\Pinf_{\hpi})$ and (1) in Lemma~\ref{lem:system}. The last step comes from Lemma~\ref{lem:performance} and the proof of Theorem~\ref{thm:main}. This concludes our proof.

\subsection{Proof of Lemma~\ref{lem:performance}}
\label{proof:lem-performance}
Let $P^h$ be the system dynamics $(\rhos,\{\Pstar_t\}_{t=1}^h,\{P_t\}_{t=h+1}^{H-1})$ for all $0\leq h\leq H-1$. Then we have
\begin{align*}
J(\pit;r,\Pstar)-J(\pit;r,P)=\sum_{h=1}^{H-1}(J(\pit;r,P^h)-J(\pit;r,P^{h-1}))+(J(\pit;r,P^0)-J(\pit;r,P)).
\end{align*}

For any $h\in[H-1]$, we have
\begin{align*}
&J(\pit;r,P^h)-J(\pit;r,P^{h-1})\\
=&\EE_{(s_1,a_1,\cdots,s_{h},a_{h})\sim(\pit,\Pstar)}\Big[\sum_{s_{h+1}}\Pstar_h(s_{h+1}|s_h,a_h)\EE_{(\pit,P)}\big[r(\tau)|s_1,a_1,\cdots,s_{h+1}\big]\\
&\qquad\qquad\qquad\qquad\qquad-\sum_{s_{h+1}}P_{h}(s_{h+1}|s_h,a_h)\EE_{(\pit,P)}\big[r(\tau)|s_1,a_1,\cdots,s_{h+1}\big]\Big]\\
=&\EE_{(s_1,a_1,\cdots,s_{h},a_{h})\sim(\pit,\Pstar)}\Big[\sum_{s_{h+1}}(\Pstar_h(s_{h+1}|s_h,a_h)-P_{h}(s_{h+1}|s_h,a_h))\EE_{(\pit,P)}\big[r(\tau)|s_1,a_1,\cdots,s_{h+1}\big]\Big]\\
\leq&\rmax\EE_{(s_h,a_h)\sim(\pit,\Pstar)}\Big[\big\Vert\Pstar_h(\cdot|s_h,a_h)-P_h(\cdot|s_h,a_h)\big\Vert_1\Big]\\
\leq&\rmax\sqrt{\frac{cC^2_P(\pit)\log(H\Ncal_{\Gcal_{P_h}}(1/N)/\delta)}{N}},
\end{align*}
where $\EE_{(\pit,P)}\big[\cdot|s_1,a_1,\cdots,s_{h+1}\big]$ is the distribution of the trajectory $\tau$ when executing policy $\pit$ under the transition probability $\{P_t\}_{t=h+1}^{H-1}$ while fixing the history to be $s_1,a_1,\cdots,s_{h+1}$. Here the first step utilizes the Tower property, the third and fourth step uses Cuachy-Schwartz inequality and the last step comes from Lemma~\ref{lem:system}.

For $J(\pit;r,P^0)-J(\pit;r,P)$, similarly we have
\begin{align*}
&J(\pit;r,P^0)-J(\pit;r,P)\leq\rmax\sqrt{\frac{cC^2_P(\pit)\log(H\Ncal_{\Gcal_{P_0}}(1/N)/\delta)}{N}}.
\end{align*}

Therefore we conclude that
\begin{align*}
J(\pit;r,\Pstar)-J(\pit;r,P)\leq H\rmax\sqrt{\frac{cC^2_P(\pit)\log(H\Ncal_{P}(1/N)/\delta)}{N}}.
\end{align*}

%% file: proof_action.tex
\section{Proof of Theorem~\ref{thm:action}}
\label{proof:thm-action}
We first derive the guarantee of MLE for estimating $\AS$. Similar to Lemma~\ref{lem:mle} and Lemma~\ref{lem:tv}, we have the following lemma in the action-based comparison setting:
\begin{lemma}
\label{lem:action}
Under Assumption~\ref{ass:realize3}, with probability at least $1-\delta$, the following event holds true:
\begin{align*}
\EE_{s\sim\mu_h,a^0\sim\mu_{0,h}(\cdot|s),a^1\sim\mu_{1,h}(\cdot|s)}\bigg[\Big\Vert P_{\hA_h}(\cdot|s,a^0,a^1)-P_{\AS_h}(\cdot|s,a^0,a^1)\Big\Vert_1^2\bigg]\leq\frac{c\log(H\Ncal_{\Gcal_{A_h}}(1/N)/\delta)}{N},\forall h\in[H].
\end{align*}
\end{lemma}
The proof is omitted here. Let $\Ecal_4$ denote the event in Lemma~\ref{lem:action}. Then under Assumption~\ref{ass:bound2}, we can apply the mean value theorem and obtain that under $\Ecal_4$, we have for all $h\in[H]$ that 
\begin{align}
&\EE_{s\sim\mu_h,a^0\sim\mu_{0,h}(\cdot|s),a^1\sim\mu_{1,h}(\cdot|s)}\bigg[|\AS_h(s,a^0)-\AS_h(s,a^1)-\hA_h(s,a^0)+\hA_h(s,a^1)|^2\bigg]\notag\\
&\qquad\leq\frac{c\kappa^2\log(H\Ncal_{\Gcal_{A_h}}(1/N)/\delta)}{N},\forall h\in[H].\label{eq:action-1}
\end{align}
Recall that $\kappa=\frac{1}{\inf_{x\in[-\rmax,\rmax]}\Phi'(x)}$.

On the other hand, note that we have the following performance lemma:
\begin{lemma}
\label{lem:performance2}
For any deterministic Markovian policies $\pi$ and $\pi'$, we have
\begin{align*}
J(\pi;\rstar,\Pstar)-J(\pi';\rstar,\Pstar)=\sum_{h=1}^H\EE_{s\sim d^{\pi'}_h}\Big[Q^{\pi}_h(s,\pi(s))-Q^{\pi}_h(s,\pi'(s))\Big]
\end{align*}
\end{lemma}
The proof is deferred to Appendix~\ref{proof:lem-performance2}. 

The rest of the proof largely follows \citet{uehara2023refined}. Under the event $\Ecal_{4}$, we can bound the suboptimality of $\hpi$ as follows:
\begin{align*}
& J(\pis;\rstar,\Pstar)-J(\hpi; \rstar, \Pstar)\leq \rmax\sum_{h=1}^H\EE_{s\sim d^{\pis}_h}\Big[\idx(\pis_h(s)\neq\hpi_h(s))\cdot\idx(\QS_h(s,\hpi_h(s))<\QS_h(s,\pis_h(s)))\Big]\\
\leq & \rmax\sum_{h=1}^H\EE_{s\sim d^{\pis}_h}\bigg[\sum_{a\in\Acal}\idx\Big(\hA_h(s,a)\geq\hA_h(s,\pis_h(s))\Big)\cdot\idx\Big(\QS_h(s,a)<\QS_h(s,\pis_h(s))\Big)\bigg],
\end{align*}
where the first step comes from Lemma~\ref{lem:performance2} and the second step is due to the definition of $\hpi$. Then for any $\alpha>0$, we have
\begin{align*}
&\EE_{s\sim d^{\pis}_h}\bigg[\sum_{a\in\Acal}\idx\Big(\hA_h(s,a)\geq\hA_h(s,\pis_h(s))\Big)\cdot\idx\Big(\QS_h(s,a)<\QS_h(s,\pis_h(s))\Big)\bigg]\\
\leq&\EE_{s\sim d^{\pis}_h}\bigg[\sum_{a\in\Acal}\idx\Big(\QS_h(s,\pis_h(s))>\QS_h(s,a)\geq\QS_h(s,\pis_h(s))-\alpha\Big)\bigg]\\
&+\EE_{s\sim d^{\pis}_h}\bigg[\sum_{a\in\Acal}\idx\Big(\QS_h(s,\pis_h(s))-\QS_h(s,a)-\hA_h(s,\pis_h(s))+\hA_h(s,a)\geq\alpha\Big)\bigg].
\end{align*}

By Assumption~\ref{ass:soft}, we have
\begin{align*}
\EE_{s\sim d^{\pis}_h}\bigg[\sum_{a\in\Acal}\idx\Big(\QS_h(s,\pis_h(s))>\QS_h(s,a)\geq\QS_h(s,\pis_h(s))-\alpha\Big)\bigg]\leq|\Acal|(\alpha/\alpha_0)^{\beta}.
\end{align*}

For the second term, we have
\begin{align*}
&\EE_{s\sim d^{\pis}_h}\bigg[\sum_{a\in\Acal}\idx\Big(\QS_h(s,\pis_h(s))-\QS_h(s,a)-\hA_h(s,\pis_h(s))+\hA_h(s,a)\geq\alpha\Big)\bigg]\\
=&\frac{1}{\alpha^2}\EE_{s\sim d^{\pis}_h}\bigg[\sum_{a\in\Acal}\alpha^2\idx\Big(\AS_h(s,\pis_h(s))-\AS_h(s,a)-\hA_h(s,\pis_h(s))+\hA_h(s,a)\geq\alpha\Big)\bigg]\\
\leq&\frac{1}{\alpha^2}\EE_{s\sim d^{\pis}_h}\bigg[\sum_{a\in\Acal}\Big|\AS_h(s,\pis_h(s))-\AS_h(s,a)-\hA_h(s,\pis_h(s))+\hA_h(s,a)\Big|^2\bigg]\\
\leq&\frac{c|\Acal|\Cact\kappa^2\log(H\Ncal_{\Gcal_{A_h}}(1/N)/\delta)}{\alpha^2N},
\end{align*}
where the last step comes from the definition of $\Cact$ and \eqref{eq:action-1}.

Therefore by picking appropriate $\alpha$, we have with probability at least $1-\delta$ that
\begin{align*}
J(\pis;\rstar,\Pstar)-J(\hpi; \rstar, \Pstar)\leq cH|\Acal|\bigg(\frac{2}{\beta}\bigg)^{\frac{\beta-2}{\beta+2}}\bigg(\frac{1}{\alpha_0}\bigg)^{\frac{2\beta}{\beta+2}}\bigg(\frac{\kappa^2\Cact\log(H\Ncal_{\Gcal_{A}}(1/N)/\delta)}{N}\bigg)^{\frac{\beta}{\beta+2}}.
\end{align*}

\subsection{Proof of Lemma~\ref{lem:performance2}}
\label{proof:lem-performance2}
For any two policies $\pi$ and $\pi'$, we have that
\begin{align*}
	&J(\pi';\rstar,\Pstar)-J(\pi;\rstar,\Pstar)\\
	=&\mathbb{E}_{\pi'}\left[\rstar_1(s_1,a_1)+V^{\pi'}_2(s_2)\right]-\E_{\pi'}\left[V^{\pi}_1(s_1)\right]\\
	=&\mathbb{E}_{\pi'}\left[V^{\pi'}_2(s_2)- (V^{\pi}_1(s_1)- \rstar_1(s_1,a_1))\right]\\
	=&\mathbb{E}_{\pi'}\left[V^{\pi'}_2(s_2)- V^{\pi}_2(s_2)\right] + \mathbb{E}_{\pi'}\left[Q^{\pi}_1(s_1,a_1)- V^{r,\pi}_1(s_1)\right]\\
	=&\mathbb{E}_{\pi'}\left[V^{\pi'}_2(s_2)- V^{\pi}_2(s_2)\right] + \mathbb{E}_{\pi'}\left[\left\langle Q^{\pi}_1(s_1,\cdot), \pi'_1(\cdot|s_1)-\pi_1(\cdot|s_1)\right\rangle\right]\\
	=&\cdots=\sum_{h=1}^H\E_{\pi'}\left[\langle Q^{\pi}_h(s_h,\cdot), \pi'_h(\cdot|s)-\pi_h(\cdot|s) \rangle\right].
\end{align*}
This concludes our proof.